\documentclass[10pt,twocolumn,letterpaper]{article}

\usepackage{iccv}              % To produce the CAMERA-READY version
\usepackage{bm}
\usepackage{algorithm, algpseudocode}
\usepackage{makecell}
\usepackage{multirow}
\usepackage{colortbl}
\usepackage{caption}
\usepackage{enumitem}
\usepackage{wrapfig} % for wrapping text around figures

\definecolor{iccvblue}{rgb}{0.21,0.49,0.74}
\usepackage[pagebackref,breaklinks,colorlinks,allcolors=iccvblue]{hyperref}

\def\paperID{993} % *** Enter the Paper ID here
\def\confName{ICCV}
\def\confYear{2025}
\newcommand{\gc}{\cellcolor[gray]{0.9}}

\title{One Perturbation is Enough: On Generating Universal Adversarial Perturbations against Vision-Language Pre-training Models}

\author{
Hao Fang\thanks{Equal contribution}
\quad
Jiawei Kong$^{*}$
\quad
Wenbo Yu
\quad
Bin Chen\thanks{Corresponding Author} 
\quad
Jiawei Li \\
Hao Wu 
\quad
Shu-Tao Xia
\quad 
Ke Xu \\
Tsinghua Shenzhen International Graduate School, Tsinghua University
\quad\\
Harbin Institute of Technology, Shenzhen\quad 
\\
{\tt\small fang-h23@mails.tsinghua.edu.cn}}

\begin{document}
\maketitle
\begin{abstract}
Vision-Language Pre-training (VLP) models have exhibited unprecedented capability in many applications by taking full advantage of the learned multimodal alignment. However, previous studies have shown they are vulnerable to maliciously crafted adversarial samples. Despite recent success, these attacks are generally instance-specific and require generating perturbations for each input sample. In this paper, we reveal that VLP models are also susceptible to the instance-agnostic universal adversarial perturbation (UAP). Specifically, we design a novel Contrastive-training Perturbation Generator with Cross-modal conditions (C-PGC). In light that the pivotal multimodal alignment in VLP models is achieved via contrastive learning, we devise to turn this powerful weapon against VLP models themselves. I.e., we employ a malicious version of contrastive learning to train the proposed generator using our carefully crafted positive and negative image-text pairs. Once training is complete, the generator is able to produce universal perturbations that can essentially destroy the established alignment relationship in VLP models. Besides, C-PGC fully utilizes the characteristics of Vision-and-Language (V+L) scenarios by incorporating both unimodal and cross-modal information as effective guidance. Extensive experiments show that C-PGC successfully forces adversarial samples to move away from their original area in the VLP model's feature space, thus fundamentally enhancing attack performance across various victim models and V+L tasks.
\end{abstract}    
\section{Introduction}
\label{sec:intro}
Vision-Language Pre-training (VLP) models have recently demonstrated remarkable efficacy in a wide range of Vision-and-Language (V+L) tasks.
% \cite{hong2019learning}. 
% including image-text retrieval \cite{wang2016comprehensive} and visual grounding \cite{hong2019learning}, and visual entailment \cite{xie2019visual}. 
By self-supervised pre-training on large-scale image-text pairs, VLP models efficiently align cross-modal features and capture rich information from the aligned multimodal embeddings, thereby providing expressive representations for various applications.

\begin{figure}
  \centering
  \includegraphics[width=0.97\linewidth]{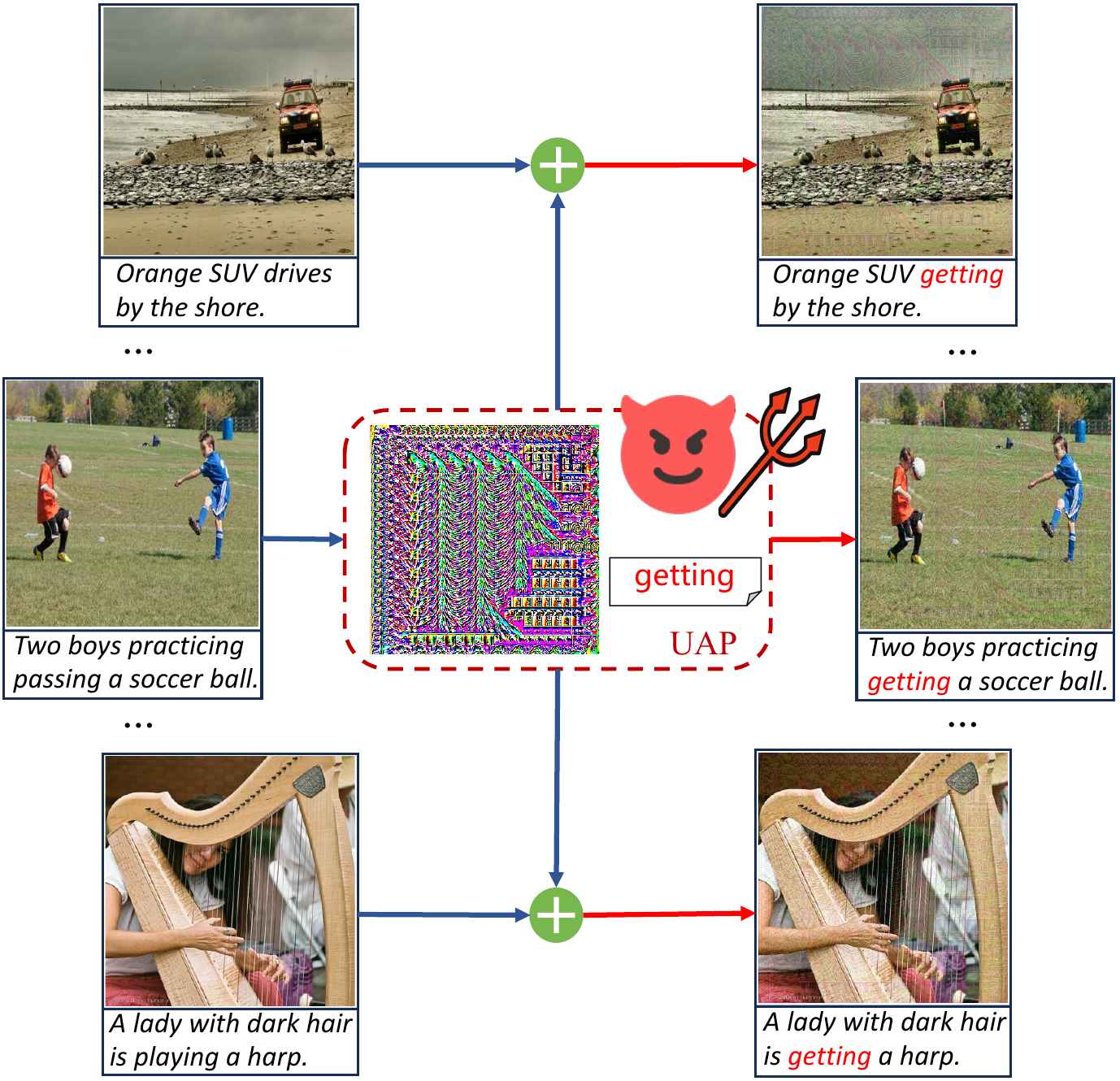}
  \centering
  \captionof{figure}{Illustration of universal adversarial attacks. With only a pair of image-text perturbations, the proposed method can effectively mislead different VLP models on diverse V+L tasks.}
\label{fig:intro_illu}
\vspace{-1em}
 \end{figure}
% \begin{figure}
% \begin{center}
% \includegraphics[width=0.7\linewidth]{imgs/introduction_p1_v5.pdf}
% \end{center}
% \vspace{-1em}
% \caption{Illustration of the universal adversarial attacks against VLP models. With only a pair of image-text perturbations, the proposed attack can effectively mislead different models on diverse V+L tasks.} 
% % which optimizes feature domains after searching the latent space.
% \vspace{-1.4em}
% \label{fig:uap}
% \end{figure}

Adversarial attacks \cite{carlini2017towards}, which aim to deceive models during inference time, have attracted extensive attention due to their significant threat to security-critical scenarios \cite{eykholt2018robust}. Recent studies have shown that VLP models are also vulnerable to adversarial samples. The pioneering work Co-Attack \cite{zhang2022towards} proposes the first multimodal attack that simultaneously perturbs both image and text modalities and displays excellent performance. However, Co-Attack only considers relatively easier white-box attacks where victim models are completely accessible. 
To handle more practical black-box settings, subsequent studies propose various transferable adversarial samples generated on an available surrogate model to fool other inaccessible models. 
Specifically, SGA \cite{lu2023set} significantly improves the adversarial transferability through the set-level cross-modal guidance obtained from data augmentations. 
Subsequently, TMM \cite{wang2024transferable} proposes to jointly destroy the modality-consistency features within the clean image-text pairs and include more modality-discrepancy features in the perturbations to further enhance transferability.
% This strategy successfully confuses VLP models and achieves outstanding transferability. 
While existing methods have achieved great success, they are all instance-specific and need to generate a perturbation for each input pair, which results in substantial computational overhead. Meanwhile, universal adversarial attacks (illustrated in Fig. \ref{fig:intro_illu}), as an efficient instance-agnostic approach that uses only one Universal Adversarial Perturbation (UAP) to conduct attacks, have not been fully investigated for VLP models.
% have exhibited remarkable performance in fooling classification tasks \cite{moosavi2017universal, poursaeed2018generative, liu2023enhancing}.
% None of these works theoretically or empirically demonstrate the vulnerability of VLP models under UAP. 
This naturally leads to a question, \emph{is it possible to design a UAP that can deceive VLP models across different image-text pairs?} 

% \begin{figure}
% \begin{center}
% \includegraphics[width=0.9\linewidth]{imgs/multiple_bar_intro.pdf}
% \end{center}
% \vspace{-1.2em}
% \caption{Performance of existing universal adversarial attacks on the text retrieval task. We transplant UAP \cite{moosavi2017universal} and GAP \cite{poursaeed2018generative} to fool VLP models. Surrogate models are ALBEF \cite{li2021align} and BLIP \cite{li2022blip}. Results of our C-PGC are presented for comparison.} 
% % which optimizes feature domains after searching the latent space.
% \vspace{-1.1em}
% \label{fig:motivation}
% \end{figure}
\begin{figure}
  \centering
  \includegraphics[width=0.93\linewidth]{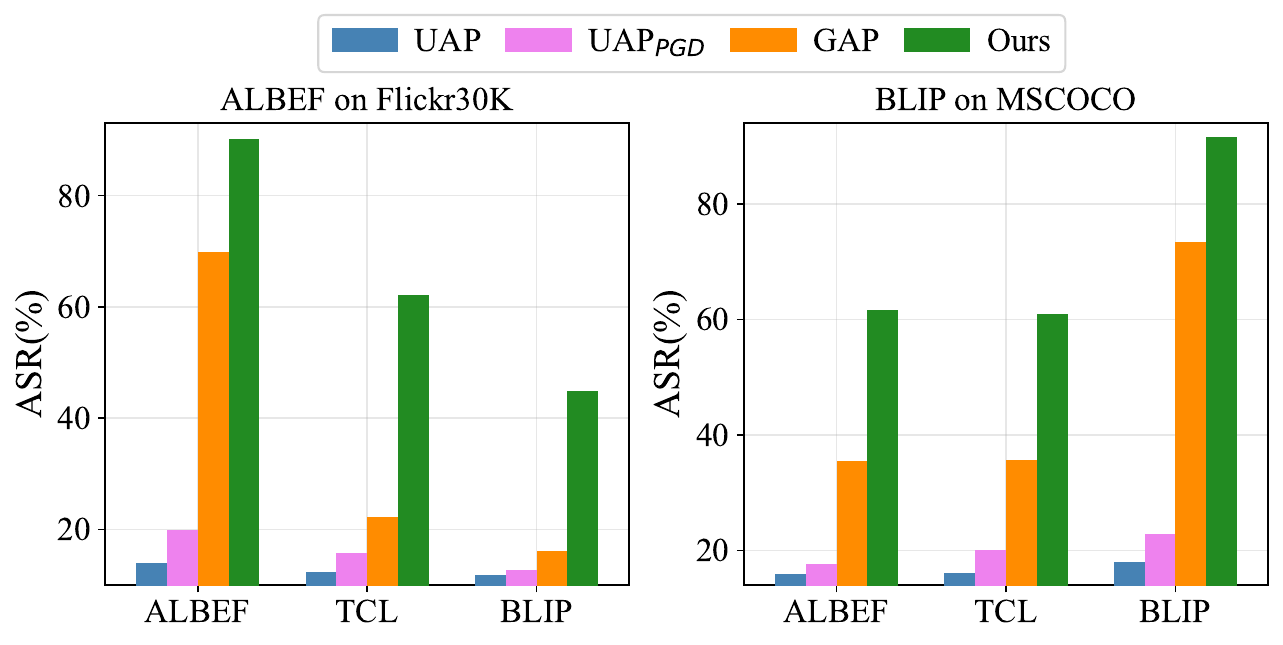}
  \centering
  \captionof{figure}{Performance of existing UAP on text retrieval with ALBEF \cite{li2021align} and BLIP \cite{li2022blip} as surrogate models. Note that UAP \cite{moosavi2017universal} is initially based on DeepFool \cite{moosavi2016deepfool} and the corresponding PGD-learned version UAP$_{PGD}$ is provided for a fair comparison.}
\label{fig:motivation}
\vspace{-1em}
 \end{figure}
 
\textbf{Motivation.} 
To this end, 
% we conduct a deep exploration of this challenging topic. First, 
we make an intuitive attempt to transplant renowned approaches UAP \cite{moosavi2017universal} and GAP \cite{poursaeed2018generative} to attack several VLP models by 
% replacing the traditional cross-entropy loss with 
maximizing the distance between the embeddings of the adversarial image and its matched texts. 
Unfortunately, Fig. \ref{fig:motivation} demonstrates that these methods yield unsatisfactory attack success rates (ASR), especially for black-box attacks. Empirically, this failure stems from their narrow focus on the image modal, disregarding the other modality and the multimodal information that plays a pivotal role in VLP models.
To overcome this challenge, we revisit VLP models' basic training paradigm and emphasize that regardless of the downstream V+L tasks, their achieved outstanding performance is heavily reliant on the well-established multimodal alignment, which draws the embeddings of matched image-text pairs closer while distancing those of non-matched pairs.
In light of this consideration, we argue that the core of an effective universal adversarial attack is to obtain a UAP that can fundamentally destroy this learned alignment relationship to mislead VLP models into making incorrect decisions. \label{sec:rationale}
% \ie, making the embedding of the generated adversarial example keep distance from its matched counterparts as far as possible to mislead VLP models into making incorrect decisions.
Besides, Fig. \ref{fig:motivation} shows that the generator-based GAP consistently outperforms UAP methods, due to the excellent distributional modeling capability of generators. This suggests the superiority of the generative paradigm, which is also corroborated by numerous studies \cite{gao2024nuat, feng2023dynamic}.
% Hence, it could be a better choice to adopt a generator-based strategy.
% obtain adversarial examples that can 
% against VLP models 

Based on these insights, we propose a novel generative framework that learns a Contrastive-training Perturbation Generator with Cross-modal conditions (C-PGC) to launch universal attacks on VLP models. 
To essentially destroy the multimodal alignment, we devise to utilize VLP models' most powerful weapons to attack against themselves. \Ie, utilize the contrastive learning mechanism to train the generator based on maliciously constructed image-text pairs that completely violate the correct V+L matching relationship, to produce perturbation that pushes the embeddings of matched pairs apart while pulling those of non-matched ones together. 
Moreover, most previous studies \cite{zhang2022towards, lu2023set, zhang2024universal} simply exploit the crucial cross-modal information by maximizing the feature distance between samples of different modals to optimize perturbations, without deeper exploration for attack enhancement. In contrast, we fully harness V+L characteristics by refining the generator architecture to incorporate cross-modal knowledge through cross-attention mechanisms for better guidance. 
Besides, we also consider the intra-modal influence and introduce a unimodal distance loss to further improve attack effectiveness. 
% Inspired by the significant improvements achieved via multimodal perturbation \cite{zhang2022towards, lu2023set}, we generate UAP for both images and texts to benefit from the synergy between different modalities.
We highlight that the proposed framework is seamlessly compatible with text perturbation generation, achieving a truly multimodal universal attack that benefits from the synergy between V+L modalities.
Our contributions are as follows:
% To the best of our knowledge, we are the first to exploit both image and text modalities to conduct UAP attacks against VLP models. 
\begin{itemize}[leftmargin=*]
\item We propose C-PGC, a novel perturbation generator conditioned on cross-modal knowledge, to produce both image and text UAPs for powerful attacks on VLP models.
% As shown in Fig. \ref{fig:intro_illu}, we demonstrate that even with only one universal perturbation, an adversary can induce serious performance degradation in the attacked VLP models.
\item We design a malicious contrastive learning paradigm that incorporates both unimodal and multimodal guidance to train the generator to produce UAP that can essentially disrupt the multimodal alignment in VLP models. 
% with meticulously constructed positive and negative pairs.
% rich information from 
% effectively undermines the modal alignment of VLP model, and greatly enhances the attack performance.
\item Extensive experiments on 6 various VLP models across diverse V+L tasks reveal that our method achieves outstanding attack performance in different scenarios. 
% (Sec. \ref{subsec:attack}, \ref{subsec:more}). C-PGC exhibits impressive transferability to multiple mainstream LVLMs (Sec. \ref{subsec:more}) and robustness under 8 considered defenses (Sec. \ref{sec:defenses}).
% \item The proposed work holds strong practical significance due to its exceptional attack effects across diverse settings, which renders it a reliable tool for evaluating the adversarial robustness of VLP models.
\end{itemize}

\section{Related Work}
\label{sec:related_work}
\subsection{Vision-Language Pre-training Models}
VLP models are pre-trained on massive image-text pairs to learn the semantic correlations across modalities and serve diverse multimodal user demands \citep{chen2023vlp, du2022survey}. We next illustrate the basis of VLP models from multiple perspectives.

% \textbf{Extraction and Representation of Features.} As for text feature extraction, similar to well-known pre-trained language models (\eg, BERT \citep{kenton2019bert}), VLP models first split the input texts into a series of subwords with a start-of-sequence token [SOS] and an end-of-sequence token [EOS] respectively inserted to the beginning and the end \citep{li2019visualbert, zhang2021vinvl, zeng2022multi}. Then, the textual representations are computed by combining the word embeddings, the positional embeddings, and the type embeddings. As for image feature extraction, VLP models exploit different types of neural networks to obtain different levels of features. Specifically, some propose to extract region-level features by pre-trained object detectors \citep{lu2019vilbert, li2019visualbert, li2020oscar}, some propose to extract grid-level features via convolutional neural networks (CNNs) \citep{li2020unicoder, wang2021simvlm}, and some others propose to extract patch-level features through Vision Transformer (ViT) \citep{dosovitskiy2020image, radford2021learning}.

\textbf{Architectures.} Based on the ways of multimodal fusion, the architectures of VLP models can be classified into two types: \emph{single-stream} and \emph{dual-stream} architectures. Single-stream architectures \citep{li2019visualbert, chen2020uniter} directly concatenate the text and image features, and calculate the attention in the same Transformer block for multimodal fusion. On the contrary, dual-stream architectures \citep{radford2021learning, li2022blip} separately feed the text and image features to different Transformer blocks and leverage the cross-attention mechanism for multimodal fusion. 
Generally, single-stream architectures are more parameter-efficient than dual-stream architectures since they adopt the same set of parameters in a Transformer block for the text and image modalities.
% Besides, the architectures of VLP models can also be divided into \emph{encoder-only architectures} and \emph{encoder-decoder architectures}. Encoder-only architectures directly send the multimodal representations output by the encoder to the final head layer, while encoder-decoder architectures first send the multimodal representations to a decoder before feeding them into the output layer.

\textbf{Pre-training Objectives.} The pre-training objectives for VLP models mainly include \emph{masked feature completion}, \emph{multimodal feature matching}, and \emph{specific downstream objectives}. Masked feature completion \citep{chen2020uniter} encourages VLP models to predict the deliberately masked tokens based on the remaining unmasked tokens.
% which improves the understanding of multimodal data for VLP models. 
% To tackle the high-dimensional and continuous characteristics of visual data, VLP models adopt two variants for masked vision tokens prediction, \ie, masked features regression \citep{tan2019lxmert, li2021unimo, xue2021probing} and masked features classification \citep{lu2019vilbert, chen2020uniter, huang2021seeing}. 
Multimodal feature matching \citep{li2021align} pre-trains VLP models to precisely predict whether the given image-text pairs are matched.
% In this way, the textual and vision features can be fully aligned by the VLP models. 
Specific downstream objectives \citep{anderson2018bottom} directly utilize the training objectives of downstream tasks (\eg, visual question answering) for pre-training VLP models. 
% By doing so, VLP models can better adapt to various downstream tasks.

\textbf{Downstream Tasks.} In this paper, we mainly consider the following multimodal downstream tasks: (1) Image-text retrieval (ITR) \citep{wang2016comprehensive}: finding the most matched image for the given text and vice versa, including image-to-text retrieval (TR) and text-to-image retrieval (IR). (2) Image caption (IC) \citep{bai2018survey}: generating the most suitable descriptions for the given image. (3) Visual grounding (VG) \citep{hong2019learning}: locating specific regions in the image that correspond with the given textual descriptions. (4) Visual entailment (VE) \citep{xie2019visual}: analyzing the input image and text and predicting whether their relationship is entailment, neutral, or contradiction.
% \subsection{VLP Models}
\subsection{Adversarial Attacks}
\textbf{Instance-specific Attacks on VLP Models.}
The adversarial robustness of VLP Models has already become a research focus. Early works \citep{kim2019single, yang2021defending} impose perturbations only on single modality and lack cross-modal interactions when attacking multimodal models. To address this issue, Co-Attack \citep{zhang2022towards} conducts the first multimodal white-box attacks on VLP models. 
% They comprehensively consider the consistency between different modalities and induce the image-text perturbations towards a stronger attack. 
On the basis of Co-Attack, \cite{lu2023set} extends the attacks to more rigorous black-box settings and proposes SGA, which utilizes set-level alignment-preserving argumentations with carefully designed cross-modal guidance. 
% By enriching the aligned image-text pairs, SGA significantly boosts the black-box adversarial transferability across different VLP models. 
% the simple interaction mechanism of 
However, \cite{wang2024transferable} points out that SGA fails to fully exploit modality correlation, and proposes TMM to better leverage cross-modal interactions by tailoring both the modality-consistency and modality-discrepancy features. Nonetheless, these methods are all instance-specific and need to craft perturbations for each input pair.
% The adversarial attacks mentioned above, whether transferable or not, are all instance-specific attacks. Instance-specific attacks redesign the perturbations for each pair of input samples. In this paper, we take a step forward and manage to generate instance-agnostic UAPs for VLP models.

\textbf{Universal Adversarial Examples}. Universal adversarial attacks \citep{moosavi2017universal, mopuri2018generalizable, zhou2023downstream} aim to deceive the victim model by exerting a uniform adversarial perturbation to all samples. 
% By adopting the paradigm of attacking all the clean samples with one uniform modification, 
These attacks save the redundant procedures of redesigning perturbations for each input sample and are hence more efficient than instance-specific methods. 
Generally, universal adversarial attacks can be categorized into optimization-based \citep{moosavi2017universal, wang2023improving, liu2023enhancing} and generation-based \citep{hayes2018learning, gao2024nuat, anil2024generating, zhou2023downstream} methods. Benefiting from the powerful modeling abilities of generative models, generation-based methods are more versatile and can produce more natural samples than optimization-based ones. A concurrent work ETU \cite{zhang2024universal} also investigates UAP on VLP models and proposes a data augmentation named ScMix. However, \textit{ETU adopts a non-generative approach that narrowly focuses on image UAP, failing to constitute a truly multimodal attack for VLP models.} Moreover, ETU demonstrates insufficient attack effects, especially for black-box transferability.
In contrast, we propose a generative multimodal attack framework based on a malicious variant of contrastive learning, which yields UAP with strong attack effects and high transferability.
\section{Universal Multimodal Attacks}
In this section, we first present the problem statement of universal adversarial attacks on VLP models. Next, we introduce the overview of our framework. Finally, we illustrate the detailed design of the proposed C-PGC.
% of each proposed technique and summarize the training objective and paradigm of C-PGC. 
% Finally, we summarize the training objective and paradigm of C-PGC.

%%%%%%%%%%%%%%%%%%%%%%%%%%%%%%%%%
\begin{figure*}[t]
\begin{center}
\includegraphics[width=0.95\linewidth]{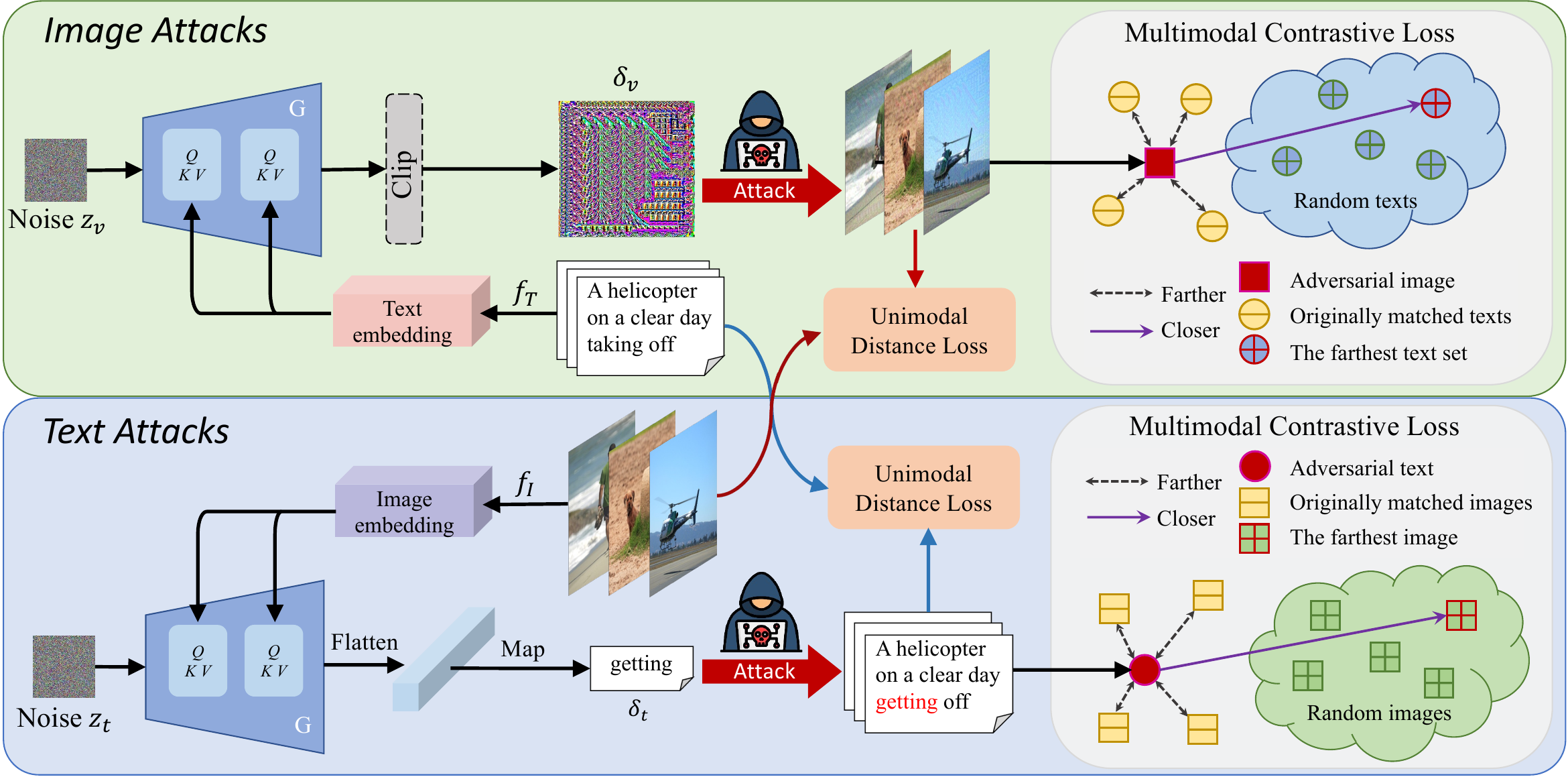}
\end{center}
\vspace{-1em}
\caption{An overview of our proposed universal adversarial attack. Benefiting from the well-designed unimodal distance loss $\mathcal{L}_{Dis}$ and multimodal contrastive loss $\mathcal{L}_{CL}$, the generator $G_w(\cdot)$, conditioned with cross-modal embeddings, learns rich knowledge from features of different modalities and thus produces $\delta_{v}$ and $\delta_{t}$ of superior generalization ability across diverse models and downstream tasks.}
\label{fig:pipeline}
\end{figure*}
%%%%%%%%%%%%%%%%%%%%%%%%%%%%%

\subsection{Problem Statement}
We define an input image-text pair as $(v, t)$ and denote $\bm e_v$ and $\bm e_t$ as the image and text embeddings encoded by the image encoder $f_{I}(\cdot)$ and text encoder $f_{T}(\cdot)$ of the targeted VLP model $f(\cdot)$. Let $\mathcal{D}_{s}$ be an available dataset consisting of image-text pairs collected by a malicious adversary. The attack objective is to utilize $\mathcal{D}_{s}$ to train a generator $G_{w}(\cdot)$ for producing a powerful pair of universal image-text perturbations $(\delta_v, \delta_t)$ that can affect the vast majority of test dataset $\mathcal{D}_{t}$ to fool models into making incorrect decisions. Formally, the attack goal can be formulated as:
\begin{equation}
\begin{split}
\mathcal{T}(f(v + \delta_v, t\oplus\delta_t)) \neq y,\; \text{s.t.}\ \|\delta_v\|_{\infty} \leq \epsilon_{v}, \|\delta_t\|_{0} \leq \epsilon_{t},
% Attention(Q,K,V)&=\text{softmax}\left(\frac{QK^{T}}{\sqrt{d}}\right) \cdot V,  
\end{split}
\label{eq:attack_goal}
\end{equation}
% Multimodal universal adversarial attacks aim to generate a pair of image-text perturbations $(\delta_v, \delta_t)$ that can affect the vast majority of samples from the target dataset to fool models into making incorrect decisions for downstream V+L tasks. 
where $\mathcal{T}(\cdot)$ denotes the operation that uses the output V+L features to obtain the final predictions, $\oplus$ indicates the text perturbation strategy  \citep{zhang2022towards, lu2023set} that replaces certain important tokens of the original sentence with crafted adversarial words, and $y$ is the correct prediction of the considered V+L task.
To ensure the perturbation's imperceptibility, we constrain the pixel-level image perturbation with $l_{\infty}$ norm of a given budget $\epsilon_{v}$.
Following previous studies \cite{zhang2022towards, lu2023set, wang2024transferable}, the textual perturbation is token-level and the stealthiness is accordingly constrained by the number of modified words $\epsilon_{t}$. To ensure the stealthiness of text perturbation, we apply a rigorous restriction that permits only a single word to be substituted ($\epsilon_{t} = 1$). On the premise of imperceptibility, the attacker attempts to generalize the crafted UAP to a wider range of test data and victim models.

% Similar to previous UAP algorithms \citep{moosavi2017universal, poursaeed2018generative}, we assume that attackers possess a certain quantity of samples from an available image-text dataset $\mathcal{D}_{t}$ to learn the universal perturbation. 
% Nevertheless, it is assumed that they do not have any prior knowledge of the test data that are used to add the perturbation and attack the VLP models. 
% The evaluated samples may either adhere to the identical (same-domain scenarios) or a different distribution (cross-domain scenarios) from the distribution of UAP's training samples. 
% Let $\mathcal{D}_{t}$ be an available dataset consisting of image-text dataset pairs collected by a malicious adversary, the attack objective is to obtain a generator $G(\cdot)$ trained on $\mathcal{D}_{t}$ to produce the UAP that can generalize to a wider range of data and victim models.

\subsection{Overview of the Proposed Framework}
% To effectively disrupt the alignment relationship, 
The overview of C-PGC is depicted in Fig. \ref{fig:pipeline}. We adopt a multimodal perturbation strategy and generate perturbations for both image and text modalities. 
Given the high similarity between the workflows of image and text, we then take image attacks as an example for illustration.

Firstly, a fixed noise $z_{v}$ is randomly initialized and subsequently fed into the conditional generator. 
For each image $v$ and its descriptions $\mathbf{t}$, the generator $G_{w}(\cdot)$ translates the input noise $z_{v}$ into the adversarial perturbation $\delta_{v}$ that is of the same size as $v$. During generation, the network $G_{w}$ additionally benefits from cross-modal information by integrating the embedding of text descriptions corresponding to the current input image $v$, \ie, $\delta_{v}=G_w(z_v;f_T(\mathbf{t}))$.
Next, the generated adversarial noise $\delta_{v}$ is injected into the clean image to obtain the adversarial image via $v_{adv} = v + \delta_{v}$. 
To better guide the training process, we design two effective unimodal and multimodal losses as our optimization objectives. 
% TODO
Unimodal loss is straightforward and aims to push the adversarial images away from the clean images in the latent embedding space, while multimodal loss is based on contrastive learning using our manually constructed positive and negative samples to strongly destroy the image-text matching relationship achieved by feature alignment. 
Once we finish training C-PGC using the proposed loss function, the input fixed noise is transformed into a UAP with great generalization and transferability.

\subsection{Detailed Design of C-PGC}
Next, we provide a detailed introduction to each of the proposed designs. Note that we primarily discuss the image attack as an example, given that the design of the text attack is completely symmetrical. The pseudocode of the training procedure is provided in Appendix A.

\textbf{Perturbation Generator Conditioned on Cross-modal knowledge.} Previous generative universal attacks \citep{gao2024nuat, anil2024generating} have shown excellent efficacy in fooling the discriminative models. Nevertheless, since existing generative attacks are limited to a single modality, directly utilizing the off-the-shelf generators might fail to leverage the multimodal interactions in these special V+L scenarios.
To address this limitation, we additionally introduce cross-modal embeddings as auxiliary information to further facilitate the process of perturbation generation. Specifically, we modify the architecture of existing decoder-based generators by adding several cross-attention modules that have been proven effective in tasks with multiple input modalities. 
The obtained textual embeddings $\bm e_{t}$ encoded by $f_{T}(\cdot)$ are then incorporated into our generator through:
\begin{equation}
\begin{split}
Q=\bm h_{t}W_{q}, K=&\ \bm e_{t}W_{k}, V = \bm e_{t}W_{v}, \\
Attention(Q,K,V)&=\text{softmax}\left(\frac{QK^{T}}{\sqrt{d}}\right) \cdot V,  
\end{split}
\label{eq:cross_attn}
\end{equation}
\noindent where $\bm{h}_t \in \mathbb{R}^{B\times d_{\alpha}}$ is the flattened intermediate features within $G_{w}(\cdot)$, and $W_{q}\in \mathbb{R}^{d_{\alpha}\times d}$, $W_{k}\in \mathbb{R}^{512\times d}$, $W_{v}\in \mathbb{R}^{512\times d}$ are the optimized parameters of attention modules.

\textbf{Multimodal Contrastive Loss.} The preceding analysis regarding the failures of existing UAP attacks encourages us to design a loss function that can guide the generated UAP to break the learned multimodal feature alignment. 
Motivated by the fact that contrastive learning underpins the cross-modal alignment, we advocate leveraging this mechanism to attack VLP models themselves by contrastively training our C-PGC to essentially disrupt the benign alignment relationship.
Concretely, we adopt the widely recognized InfoNCE \citep{he2020momentum} as our basic contrastive loss.

To establish the contrastive paradigm, we first define the adversarial image $v_{adv}$ as the anchor sample. Besides, it is also necessary to construct an appropriate set of positive and negative samples.
Based on the fundamental objective of our attack, it is natural to leverage the originally matched text description set \(\mathbf{t}=\{t_1, t_2, \dots, t_M\}\) as negative samples \(\mathbf{t}_{neg}\) to amplify the discrepancy between matched image-text pairs in the feature space of VLP models. Additionally, we need to select a set of positive samples to further pull the adversarial image \(v_{adv}\) away from its corresponding text descriptions \(\mathbf{t}\). 
To this end, we propose a \textit{farthest selection strategy}, which associates the anchor image \( v_{adv} \) with target texts $\mathbf{t}_{pos}$ whose embeddings differ significantly from that of the original clean image $v$, to reach a more strong disruption of the multimodal alignment.
Specifically, we randomly sample a batch of text sets from $\mathcal{D}_{s}$ and select the text set with the largest feature distances from the current image $v$ as positive samples, \ie, $\mathbf{t}_{pos} = \{t_1', t_2'\dots, t_K'\}$.
Moreover, we utilize data augmentations that resize the clean $v$ into diverse scales and apply random Gaussian noise to acquire a more diverse image set $\mathbf{v} = \{v_1, v_2\dots, v_N\}$ for set-level guidance \cite{lu2023set}. With these well-constructed positive and negative samples, the multimodal contrastive loss $\mathcal{L}_{CL}$ can be formulated as:
\begin{equation}
% \mathcal{L}_{CL} = log\Bigg(\frac{{\underset{i=1}{\overset{M}{\sum}}}{\underset{j=1}{\overset{S}{\sum}}} exp(S(t_i, v_j) / \tau)}{ {\underset{i=1}{\overset{M}{\sum}}}{\underset{j=1}{\overset{S}{\sum}}} exp(S(t_i, v_j) / \tau) +{\underset{i=1}{\overset{K}{\sum}}}{\underset{j=1}{\overset{S}{\sum}}} exp(S(t_i', v_j) / \tau)}\Bigg)  
% {\underset{i=1}{\overset{n}{\sum}}}
% \mathcal{L}_{CL} = log\Bigg(\frac{{\underset{i=1}{\overset{M}{\sum}}}{\underset{j=1}{\overset{S}{\sum}}} exp(S(t_i, v_j) / \tau)}{ {\underset{j=1}{\overset{S}{\sum}}} \big( {\underset{i=1}{\overset{M}{\sum}}} exp(S(t_i, v_j) / \tau) +{\underset{i=1}{\overset{K}{\sum}}} exp(S(t_i', v_j) / \tau)\big)}\Bigg)  
% d(v, t) = e^{Sim(v, t) / \tau}, \\
% \mathcal{L}_{CL} = log\Big(\frac{\sum_{i=1}^{S}\sum_{j=1}^{M} d(v_i, t_j)}{\sum_{i=1}^{S}\sum_{j=1}^{M} d(v_i, t_j) + {\sum_{i=1}^{S}\sum_{j=1}^{K} d(v_j, t_j') }}\Big)
% \mathcal{L}_{CL} = log\Bigg(\frac{{\underset{i=1}{\overset{S}{\sum}}}{\underset{j=1}{\overset{M}{\sum}}} e^{d(v_i+\delta_v, t_j)/\tau}}{{\underset{i=1}{\overset{S}{\sum}}}{\underset{j=1}{\overset{M}{\sum}}} e^{d(v_i+\delta_v, t_j)/\tau} + {\underset{i=1}{\overset{S}{\sum}}}{\underset{j=1}{\overset{K}{\sum}}} e^{d(v_i+\delta_v, t_j')/\tau} }\Bigg), 
\mathcal{L}_{CL} = \log\frac{{\underset{i=1}{\overset{N}{\sum}}}{\underset{j=1}{\overset{M}{\sum}}} s(v_i+\delta_v, t_j)}{{\underset{i=1}{\overset{N}{\sum}}}{\underset{j=1}{\overset{M}{\sum}}} s(v_i+\delta_v, t_j) + {\underset{i=1}{\overset{N}{\sum}}}{\underset{j=1}{\overset{K}{\sum}}} s(v_i+\delta_v, t_j') }, 
\label{eq:infoNCE}
\end{equation}
\noindent where $\mathbf{\delta}_{v}$ is the universal image perturbation; $s(v, t) = \exp(\text{sim}(f_I(v), f_T(t))/\tau)$, where $\tau$ denotes the temperature parameter and $\text{sim}(\cdot, \cdot)$ represents the cosine similarity.

\textbf{Unimodal Distance Loss.} Apart from the multimodal guidance, we also consider the unimodal influence by directly pushing adversarial images away from their initial visual semantic area to improve the attack.
% Similarly, we apply separate augmentations to the image $v$ to obtain several views of $v$ and then construct positive pairs using $v$ and its augmented images. 
% Inspired by this operation, an adversary can maliciously leverage a similar method to further promote the attack effects. 
Similarly, the input image $v$ also undergoes resizing and noise perturbation
% resized to different scales and then added with Gaussian noise 
to generate the augmented image set $\mathbf{v}=\{v_1, v_2\dots, v_N\}$ for set-level guidance. Then, we craft the adversarial image through $v_{adv} =v + \delta_{v}$ and process $v_{adv}$ with the same augmentation operation to obtain the adversarial image set $\mathbf{v}_{adv} = \{v_1^{adv}, v_2^{adv}\dots, v_N^{adv}\}$. 
Finally, we minimize the negative Euclidean distance between the embeddings of adversarial images and clean images to optimize the UAP generator. Formally, the loss $\mathcal{L}_{Dis}$ is formulated as:
\begin{equation}
\mathcal{L}_{Dis} = -{\underset{i=1}{\overset{N}{\sum}}}{\underset{j=1}{\overset{N}{\sum}}}\|f_I(v_{i}^{adv}) - f_I(v_{j})\|_2.
\label{loss:dis}
\end{equation}
Taking advantage of the unimodal guidance, $\mathcal{L}_{Dis}$ ensures an effective optimization direction during the generator training and further enhances the attack effectiveness.

\textbf{Training Objective.} With the above two well-designed loss terms $\mathcal{L}_{Dis}$ and $\mathcal{L}_{CL}$, the overall optimization objective of our conditional generator for image attacks can be formulated as:
\begin{equation}
\begin{split}
\min\limits_{w} \mathbb{E}_{(v, \mathbf{t})\sim\mathcal{D}_s, \mathbf{t}_{pos} \sim \mathcal{D}_s} ( \mathcal{L}_{CL} + \lambda\mathcal{L}_{Dis}),\; \\ \text{s.t. } \|G_{w}(z_v;f_T(\mathbf{t})) \|_{\infty} \leq \epsilon_{v},
\label{eq:total}
\end{split}
\end{equation}

\noindent where $\lambda$ is the pre-defined hyperparameter to balance the contributions of $\mathcal{L}_{CL}$ and $\mathcal{L}_{Dis}$. % 如果注释掉下一行的话，需要在这句末尾把句号加上
% , which is set to $0.1$ by default and is proved to be the most effective one in our ablation study.  
By training the network with the proposed loss function over the entire data distribution of the multimodal training dataset \( D_{s} \), the generator \( G_{w}(\cdot) \) is optimized to generate UAPs that push the features of mismatched image-text pairs together while pulling the embeddings of the matched ones apart. This finally enables the generation of UAPs with strong generalization capabilities and high adversarial transferability.

\textbf{Text Modality Attacks}. \label{para:textmodal}
In textual attacks, the generator architecture and training loss are completely symmetrical with those of image attacks. Correspondingly, embeddings of the matched image $v$ are used as the cross-modal conditions for the generator. 
Given an adversarial text $t_{adv}$ as the anchor sample, we use the set $\mathbf{v}=\{v_1, v_2\dots,v_N\}$ scaled from the originally matched image $v$ as negative samples while the $\mathbf{v}'=\{v_1', v_2'\dots,v_N'\}$ augmented from the farthest image $v'$ within the randomly sampled image set as positive samples to formulate the $\mathcal{L}_{CL}$ loss. 
% To further improve the performance, we also apply augment $v$ and $v'$ into different scales to . 
$\mathcal{L}_{Dis}$ is consequently calculated as the negative Euclidean distance between the embeddings of $t_{adv}$ and the clean input $t$. Accordingly, the conditional generator is utilized to output the adversarial textual embeddings, which are subsequently mapped back to the vocabulary space to obtain a universally applicable word-level perturbation. 
 
% Detailed math expressions of $\mathcal{L}_{CL}$ and $\mathcal{L}_{Dis}$ are presented in Appendix \ref{}.

A notable distinction between image and text attacks is the way to inject adversarial perturbations. 
% Due to the discreteness of text data, w
We align with previous studies \cite{zhang2022towards, lu2023set, wang2024transferable} and apply the token-wise substitute strategy that replaces certain important words in the original sentence with crafted adversarial words. 
% Note: 如果字数不够，这里可以补充一下对输出embedding进行了裁剪。
Prior to the word replacement, a meticulous process is undertaken to identify the most optimal position within the sentence to insert the perturbation. 
Our strategy intends to replace the words that are more likely to have a greater influence during decision-making. Concretely, for each word $w_i$ within a given sentence, we compute the distance between the embeddings of the original sentence and the $w_i$-masked version to determine its contribution. 
By convention, the imperceptibility of text UAP is controlled by the number of modified words $\epsilon_{t}$ \cite{behjati2019universal}. As aforementioned, we set $\epsilon_{t}=1$ for high stealthiness, \ie, choose the single word exerting the highest feature distance as the target for replacement.
\begin{table*}[t]
  \centering
  \setlength{\tabcolsep}{5.8pt}
  \caption{ASR (\%) of different methods for image-text retrieval tasks on Flickr30k dataset. TR indicates text retrieval based on the input image, while IR is image retrieval using the input text. The results on the MSCOCO dataset are in Appendix C due to space limit.}
    \resizebox{0.93\linewidth}{!}{\begin{tabular}{cc|cc|cc|cc|cc|cc|cc} \toprule
    \multirow{2}[0]{*}{Source} & \multirow{2}[0]{*}{Method} & \multicolumn{2}{c}{ALBEF} & \multicolumn{2}{c}{TCL} & \multicolumn{2}{c}{X-VLM} & \multicolumn{2}{c}{CLIP$_\textrm{ViT}$} & \multicolumn{2}{c}{CLIP$_\textrm{CNN}$} & \multicolumn{2}{c}{BLIP} \\ \cmidrule(lr){3-14} 
          &        & TR   & IR   & TR   & IR   & TR   & IR   & TR   & IR   & TR   & IR   & TR   & IR \\ \midrule
         \multirow{3}[0]{*}{ALBEF} & GAP   & \gc 69.78    & \gc 81.59    & 22.15    & 29.97    & 6.61     & 18.37    & 23.40     & 37.54    & 29.92    & 44.29    & 16.09    & 28.12 \\ & ETU & \gc78.01    & \gc84.56    & 29.92    & 35.91    & 14.33    & 22.03    & 23.77    & 39.20     & 33.55    & 47.69    & 22.61    & 32.28  \\
            & Ours  & \gc\textbf{90.13} & \gc\textbf{88.82} & \textbf{62.11} & \textbf{64.48} & \textbf{20.53} & \textbf{39.38} & \textbf{43.10} & \textbf{65.93} & \textbf{54.40} & \textbf{72.51} & \textbf{44.79} & \textbf{56.36}  \\ \midrule
          \multirow{3}[0]{*}{TCL} & GAP   & 33.50     & 40.61    & \gc82.41    & \gc80.67    & 6.61     & 17.79    & 21.55    & 38.56    & 30.57    & 45.48    & 21.45    & 31.82 \\ & ETU      & 28.26    & 35.03    & \gc90.48    & \gc87.57    & 9.65     & 20.56    & 25.00       & 39.68    & 36.14    & 49.33    & 18.93    & 29.19 \\
                 & Ours  & \textbf{50.26} & \textbf{56.29} & \gc\textbf{94.93} & \gc\textbf{90.64} & \textbf{14.94} & \textbf{33.96} & \textbf{46.92} & \textbf{66.41} & \textbf{52.98} & \textbf{70.66} & \textbf{35.75} & \textbf{52.52} \\   \midrule
           \multirow{3}[0]{*}{X-VLM} & GAP   & 16.14    & 24.43    & 17.08    & 26.20     & \gc90.24    & \gc85.98    & 24.51    & 41.15    & 42.62    & 53.08    & 16.19    & 25.74 \\ & ETU      & 12.33    & 21.93    & 13.98    & 24.04    & \gc93.19    & \gc90.85    & 23.89    & 39.62    & 35.62    & 51.19    & 12.09    & 23.59 \\
            & Ours  & \textbf{24.46} & \textbf{47.77} & \textbf{29.19} & \textbf{50.15} & \gc\textbf{93.29} & \gc\textbf{91.90} & \textbf{43.47} & \textbf{66.03} & \textbf{59.20} & \textbf{72.79} & \textbf{32.39} & \textbf{52.24} \\ \midrule
           \multirow{3}[0]{*}{CLIP$_\textrm{ViT}$} & GAP   & 11.72    & 23.34    & 15.32    & 26.39    & 8.54     & 20.48    & \gc85.73    & \gc90.45    & 48.83    & 60.78    & 14.83    & 26.46 \\ & ETU      & 14.80     & 25.23    & 21.22    & 30.87    & 10.87    & 24.96    & \gc84.14    & \gc90.45    & 57.51    & 65.51    & 16.40     & 27.22 \\
                & Ours  & \textbf{23.23} & \textbf{38.67} & \textbf{25.05} & \textbf{41.79} & \textbf{15.85} & \textbf{35.59} & \gc\textbf{88.92} & \gc\textbf{93.05} & \textbf{66.06} & \textbf{75.42} & \textbf{26.71} & \textbf{45.70} \\  \midrule
           \multirow{3}[0]{*}{CLIP$_\textrm{CNN}$} & GAP   & 13.57    & 25.21    & 19.05    & 28.87    & 11.59    & 23.13    & 27.46    & 43.16    & \gc73.18    & \gc81.60     & 15.25    & 27.94 \\ & ETU      & 8.94     & 20.59    & 13.25    & 24.41    & 8.94     & 20.82    & 21.92    & 40.51    & \gc91.71    & \gc92.40     & 11.15    & 23.82 \\
                 & Ours     & \textbf{19.01} & \textbf{41.86} & \textbf{22.98} & \textbf{47.02} & \textbf{19.61} & \textbf{43.26} & \textbf{40.89} & \textbf{65.77} & \gc\textbf{96.50} & \gc\textbf{94.22} & \textbf{24.19} & \textbf{48.17} \\ \midrule
           \multirow{3}[0]{*}{BLIP} & GAP    & 12.23    & 23.94    & 14.49    & 25.44    & 6.91     & 17.81    & 20.32    & 37.00       & 26.81    & 43.59    & \gc47.21    & \gc73.33 \\ & ETU      & 19.32    & 27.91    & 19.98    & 29.15    & 11.99    & 20.91    & 24.38    & 39.84    & 31.61    & 46.22    & \gc59.52    & \gc77.82 \\
            & Ours  & \textbf{32.17} & \textbf{44.40} & \textbf{33.44} & \textbf{44.51} & \textbf{18.60} & \textbf{35.53} & \textbf{43.35} & \textbf{60.26} & \textbf{48.96} & \textbf{66.95} & \gc\textbf{71.82} & \gc\textbf{82.82}  \\ \bottomrule
    \end{tabular} }
  \label{tab:main}%
\end{table*}%

\section{Experiments}
To comprehensively validate the effectiveness of C-PGC, we conduct experiments on diverse downstream V+L tasks across multiple VLP models. 
% Besides, sufficient ablation studies in Sec. \ref{subsec:ablation} validate the contribution of each proposed technique and explore the impact of several crucial factors.
% We first present the experimental setup in Sec. \ref{subsec:setup} and then comprehensively evaluate C-PGC across multiple VLP models in Sec. \ref{subsec:attack}. 
% Sec. \ref{subsec:more} presents results on more downstream V+L tasks to further validate the effectiveness. Besides, sufficient ablation studies in Sec. \ref{subsec:ablation} validate the contribution of each proposed technique and explore the impact of several crucial factors. 
Please see more experiments including \textbf{performance under defenses, cross-domain attacks, loss analysis, and visualization} in the Appendix.

\subsection{Experimental Setup} \label{subsec:setup}
\textbf{Downstream tasks and datasets.}
We conduct a comprehensive study of C-PGC on four downstream V+L tasks, including image-text retrieval (ITR), image captioning (IC), visual grounding (VG), and visual entailment (VE). For ITR tasks, we employ the Flickr30K \citep{plummer2015flickr30k} and MSCOCO \citep{lin2014microsoft} datasets which are commonly used in previous works \citep{zhang2022towards, lu2023set}. The MSCOCO is also adopted for evaluating the IC task. For VG and VE tasks, we evaluate on SNLI-VE \citep{xie2019visual} and RefCOCO+ \citep{yu2016modeling} datasets respectively. 
% Their used datasets are as follows: 
% \begin{itemize}
%     \item \textbf{Flickr30K \citep{plummer2015flickr30k}.} Collected from the Flickr website, this dataset describes different items and activities, which becomes a standard benchmark for various V+L tasks. It contains 31,783 images, each of which has five associated captions. We use it for ITR tasks.
%     \item \textbf{MSCOCO \citep{lin2014microsoft}.} The MSCOCO dataset is a rich and diverse dataset consisting of 123,287 images, each of which is annotated with approximately five sentences. We use this dataset to test the attack performance of ITR and IC tasks.
%     \item \textbf{SNLI-VE \citep{xie2019visual}.} Originally proposed for natural language reasoning tasks, this dataset provides large-scale images and descriptions, where each image is annotated with several sentences and their logical relationship labels, including entailment, neutral, and contradiction. This dataset is used for VE tasks.
%     \item \textbf{RefCOCO+ \citep{yu2016modeling}.} RefCOCO+ is an image dataset selected from MSCOCO. It contains 19,992 images and 141,564 annotations, which is specially used for visual grounding (VG) tasks.
% \end{itemize}

\textbf{Models.} We conduct experiments on a wide range of VLP models including ALBEF \citep{li2021align}, TCL \citep{yang2022vision}, X-VLM \citep{zeng2022multi}, CLIP$_\textrm{ViT}$ \citep{radford2021learning}, CLIP$_\textrm{CNN}$ \citep{radford2021learning}, and BLIP \citep{li2022blip}. 
% 1Different image feature extraction modules are utilized for CLIP, namely ViT-B/16 for CLIP$_\textrm{ViT}$ and ResNet-101 for CLIP$_\textrm{CNN}$. 
% while in other models only ViT is leveraged for image feature extraction. 
Note that for different V+L tasks, we correspondingly select different VLP models for evaluation based on their capability \citep{wang2024transferable}. For instance, among the six considered VLP models, only ALBEF, TCL, and X-VLM can handle VG tasks, while only ALBEF and TCL can deal with VE tasks. 

\textbf{Baselines.} We transplant the representative GAP \citep{poursaeed2018generative} to V+L scenarios by appropriately editing its original loss function \citep{lu2023set}. We also consider a concurrent UAP study ETU \cite{zhang2024universal}, which adopts a non-generative method that narrowly focuses on image perturbation, despite the multimodal nature of V+L scenarios. Note that it implements several versions and we report their best results.
% Besides, we also compare with the state-of-the-art (SOTA) instance-specific methods SGA \citep{lu2023set} and TMM \citep{wang2024transferable} To reveal our achieved great black-box transferability. However, it is important to note that universal adversarial attacks usually do not directly compare with instance-specific methods due to fairness considerations.

\textbf{Implementation details.}
Following the SGA \citep{lu2023set}, we adopt Karpathy split \citep{karpathy2015deep} to preprocess the dataset and build the test set for evaluation. The test set is disjoint with the generator's training data for rigorous assessment. To ensure perturbation invisibility, we follow \citep{wang2024transferable} and limit the perturbation budgets $\epsilon_v$ to $12/255$ and $\epsilon_t$ to $1$. For augmentation, we resize the original images into five scales $\{0.5, 0.75, 1, 1.25, 1.5\}$, and apply Gaussian noise $\mathcal{N}(0, 0.5^2)$. See Appendix B for more details.

\subsection{Universal Attack Effectiveness} \label{subsec:attack}
To align with previous studies \citep{zhang2022towards, lu2023set}, we first consider the typical V+L task image-text retrieval and calculate the ASR as the proportion of successful adversarial samples within the originally correctly predicted pairs. We present the performance based on R@1 retrieval results in Table \ref{tab:main}. Appendix C supplements the results of R@5 and R@10.
% Experimental results across six VLP models are presented in Table \ref{tab:main}.
% We also provide the visualization of the image retrieval on the MSCOCO dataset in Fig. \ref{fig:IR}.

% Table generated by Excel2LaTeX from sheet 'compare'
% \begin{table}[htbp]
% \setlength{\tabcolsep}{10pt}
%   \centering
%   \caption{ASR (\%) Comparison of our method with the SOTA instance-specific methods on ITR tasks using the Flickr30K dataset. The substitute model is ALBEF and we present the black-box fooling rates on different target models.}
%     \begin{tabular}{l|ccc} \toprule
%     \multicolumn{1}{l}{Target model} & \multicolumn{1}{l}{Method} & TR    & IR \\ \midrule
%     \multirow{3}[0]{*}{TCL} & SGA   & 43.95 & 48.83 \\
%           & TMM   & 64.97 & \textbf{69.60} \\ 
%           & Ours  & \textbf{66.67} & 69.59 \\  \midrule
%     \multirow{3}[0]{*}{XVLM} & SGA   & 26.54 & 39.11 \\ 
%           & TMM   & \textbf{47.14} & \textbf{55.49} \\
%           & Ours  & 31.40  & 44.75 \\ \midrule
%     \multirow{3}[0]{*}{CLIP$_\textrm{ViT}$} & SGA   & 33.83 & 43.57 \\ 
%           & TMM   & 52.90  & 60.90 \\
%           & Ours  & \textbf{60.84} & \textbf{73.56} \\ \midrule
%     \multirow{3}[0]{*}{CLIP$_\textrm{CNN}$} & SGA   & 34.39 & 46.68 \\ 
%           & TMM   & 56.61 & 62.97 \\
%           & Ours  & \textbf{74.22} & \textbf{81.62} \\ \midrule
%     \multirow{3}[0]{*}{BLIP} & SGA   & 39.65 & 53.65 \\
%           & TMM   & 59.99 & 66.01 \\
%           & Ours  & \textbf{72.13} & \textbf{73.84} \\  \bottomrule
%     \end{tabular}%
%   \label{tab:sga_tmm}%
% \end{table}%
\begin{table*}[t]
  \setlength{\tabcolsep}{6.8pt}
  \centering
  \caption{Performance of C-PGC on the visual grounding task. The first row displays the source models, and the Baseline indicates the clean performance of the target model on clean data.}
    \resizebox{0.88\linewidth}{!}{\begin{tabular}{c|ccc|ccc|ccc|ccc}
    \toprule
    \multirow{2}[0]{*}{Target} & \multicolumn{3}{c}{Baseline} & \multicolumn{3}{c}{ALBEF} & \multicolumn{3}{c}{TCL} & \multicolumn{3}{c}{X-VLM} \\ \cmidrule(lr){2-13}
    % \cmidrule(lr){2-4} \cmidrule(lr){5-7} \cmidrule(lr){8-10} \cmidrule(lr){11-13}
    & Val & TestA & TestB & Val & TestA & TestB & Val & TestA & TestB & Val & TestA & TestB \\
    \midrule
    ALBEF & 58.4     & 65.9     & 46.2     & \gc\textbf{37.1}     & \gc\textbf{39.8}     & \gc\textbf{32.0}     & 42.2     & 46.9     & 35.2     & 37.6     & 40.2     & 33.0 \\
    TCL & 59.6     & 66.8     & 48.1     & 43.6     & 47.8     & 36.9     & \gc\textbf{39.0}     & \gc\textbf{41.4}     & \gc\textbf{33.6}     & 39.5     & 41.7     & 34.1 \\
    X-VLM & 70.8     & 67.8     & 61.8     & 51.8     & 54.7     & 47.7     & 52.7     & 55.9     & 47.8     & \gc\textbf{33.1}     & \gc\textbf{34.7}     & \gc\textbf{28.8} \\
    \bottomrule
    \end{tabular}}
  \label{tab:vg}
\end{table*}

\textbf{White-box attack performance.} By observing the white-box ASR in the gray-shaded area, we demonstrate that the proposed algorithm stably achieves excellent ASR on all the evaluated VLP models, validating the outstanding capability of the produced UAP. 
With only a single pair of perturbations, we reach a noteworthy average white-box ASR of nearly 90\% on two large datasets in terms of both TR and IR tasks. Compared with GAP \cite{poursaeed2018generative} and ETU \cite{zhang2024universal}, the proposed method consistently enhances the fooling rates in the white-box scenario, confirming the great validity of our suggested multimodal contrastive-learning mechanism. 
% Especially on the MSCOCO dataset, our method achieves over 95\% average ASR on ITR tasks across six surrogate models.
Essentially, the exceptional performance stems from the efficacy of our generated UAP in destroying the alignment between the image and text modalities, thereby misleading the VLP model during inference.

\textbf{Black-box attack performance}. We also conduct thorough experiments regarding the adversarial transferability of the generated UAP by transferring from surrogate models to other inaccessible models. 
As demonstrated in Table \ref{tab:main}, the proposed C-PGC displays great attack effects in the more realistic black-box scenarios, \eg, 72.51\% from ALBEF to CLIP$_\textrm{CNN}$ for IR tasks. 
We highlight that the advantage of C-PGC over the concurrent study ETU is greatly amplified in the more challenging black-box scenarios, which achieves a significant average improvement of 17.76\% on the Flickr30K dataset. 
These experimental results indicate that our generative contrastive learning framework does not overly rely on the encoded feature space tailored to the surrogate model. Conversely, it is well capable of transferring to breaking the multimodal alignment of other unseen target models, thus attaining superior adversarial transferability.

\subsection{Evaluation on More Downstream Tasks} \label{subsec:more}
% To further reveal the effectiveness of the proposed algorithm, we consider a range of scenarios including more downstream V+L tasks, cross-domain transfer, and attacking the cutting-edge large vision-language models (LVLMs). 
% TODO: 如果页数不够，再用这个东西凑数，否则删除。
% Finally, we visualize the t-SNR clustering results using the embedding of generated samples and original clean samples, which confirmed the effectiveness of the attack.

% \textbf{More downstream tasks.} 
% Regardless of the downstream tasks, the cross-modal interactions and alignment play an important part in multimodal learning. 
We then provide results on more downstream V+L tasks. Specifically, we consider Image Captioning (IC), Visual Grounding (VG), and Visual Entailment (VE). The results of VE are shown in Appendix C due to space limit.

\textbf{Visual grounding.} This is another common V+L task, which aims to locate the correct position in an image based on a given textual description. We conduct experiments on RefCOCO+ using ALBEF, TCL, and X-VLM as source and target models. Table \ref{tab:vg} indicates that C-PGC brings a notable negative impact on the localization accuracy in both white-box and black-box settings, again verifying that the produced UAP strongly breaks the cross-modal alignment.
\begin{table}[htbp]
  \centering
  \caption{Performance of C-PGC on image captioning task. The Baseline represents the clean performance. The surrogate is BLIP.}
    \resizebox{0.97\linewidth}{!}{\begin{tabular}{cccccc} \toprule
    Source & B@4 & METEOR & ROUBE\_L & CIDEr & SPICE \\ \midrule
    Baseline & 39.7     & 31.0     & 60.0     & 133.3    & 23.8 \\
    ALBEF & 30.1     & 23.7     & 51.2     & 92.5     & 17.5 \\
    TCL   & 29.5     & 23.5     & 51.0     & 88.9     & 17.3 \\
   \gc BLIP  & \gc\textbf{21.2} & \gc\textbf{19.1} & \gc\textbf{45.5} & \gc\textbf{62.5} & \gc\textbf{13.7} \\ \bottomrule
    \end{tabular}}
    \vspace{-1em}
  \label{tab:caption}%
\end{table}

% \begin{wraptable}{r}{0.44\linewidth}
%   \centering
%   \caption{Performance of visual grounding. The Baseline indicates the performance of the target model on clean data.}
%     \resizebox{0.99\linewidth}{!}{\begin{tabular}{l|c|ccc} \toprule
%     Source & Target & Val   & TestA & TestB \\ \midrule
%     Baseline & \multirow{4}[0]{*}{ALBEF} & 58.4  & 65.9  & 46.2 \\
%     ALBEF &       & \textbf{43.1} & \textbf{39.0} & \textbf{32.0} \\
%     TCL   &       & 47.2  & 46.7  & 35.2 \\
%     X-VLM  &       & 44.1  & 39.3  & 32.9 \\    \midrule
%     Baseline & \multirow{4}[0]{*}{TCL} & 59.6  & 66.8  & 48.1 \\
%     ALBEF &       & 53.9  & 47.9  & 37.1 \\
%     TCL   &       & \textbf{42.1} & \textbf{41.3} & \textbf{33.3} \\
%     X-VLM  &       & 42.5  & 41.8  & 34.2 \\    \midrule
%     Baseline & \multirow{4}[0]{*}{X-VLM} & 47.6  & 53.8  & 42.2 \\
%     ALBEF &       & 40.3  & 43.1  & 35.9 \\
%     TCL   &       & 38.6  & 43.5  & 35.7 \\
%     X-VLM  &       & \textbf{34.3} & \textbf{37.7} & \textbf{32.9} \\   \bottomrule
%     \end{tabular}}
%   \label{tab:vg}
% \end{wraptable}

\textbf{Image captioning.} The objective of IC is to 
% encode an input image into image embeddings and subsequently utilize an image-conditioned language modeling module to 
generate text descriptions relevant to the semantic content based on the given image. We use ALBEF, TCL, and BLIP as source models and attack the commonly used captioning model BLIP. Similar to SGA \citep{lu2023set}, several typical evaluation metrics of IC are calculated to measure the quality of generated captions, including BLEU \citep{papineni2002bleu}, METEOR \citep{banerjee2005meteor}, ROUGE \citep{lin2004rouge}, CIDEr \citep{vedantam2015cider}, and SPICE \citep{anderson2016spice}. The results in Table \ref{tab:caption} demonstrate that our algorithm again displays prominent attack effectiveness, \eg, the crated UAP induces notable drops of 10.2\% and 9\% in the B@4 and ROUGE\_L respectively when transferred from TCL to BLIP.

\subsection{Ablation Study} \label{subsec:ablation}
This part employs ALBEF \citep{li2021align} as the surrogate model and provides ablation studies on Flickr30K. 
We begin our analysis on the contribution of each proposed technique. Next, we examine the sensitivity of certain hyperparameters.

\noindent\textbf{The effect of $\mathcal{L}_{CL}$ and $\mathcal{L}_{Dis}$.} To investigate the impact of the proposed loss terms, we introduce two variants C-PGC$_{CL}$ and C-PGC$_{Dis}$ that remove $\mathcal{L}_{CL}$ and $\mathcal{L}_{Dis}$ from the overall training loss respectively. 
\begin{table*}[t]
\setlength{\tabcolsep}{6.8pt}
  \centering
  \caption{ASR (\%) of C-PGC and its variants averaged across six target models on retrieval tasks.} \label{tab:ablation}
    \resizebox{0.9\linewidth}{!}{\begin{tabular}{l|cc|cc|cc|cc|cc|cc} \toprule
    \multirow{2}[0]{*}{Method} & \multicolumn{2}{c}{ALBEF} & \multicolumn{2}{c}{TCL} & \multicolumn{2}{c}{X-VLM} & \multicolumn{2}{c}{CLIP$_\textrm{ViT}$} & \multicolumn{2}{c}{CLIP$_\textrm{CNN}$} & \multicolumn{2}{c}{BLIP} \\ \cmidrule(lr){2-13} 
    % \cmidrule(lr){2-3} \cmidrule(lr){4-5} \cmidrule(lr){6-7} \cmidrule(lr){8-9} \cmidrule(lr){10-11} \cmidrule(lr){12-13}
         &  TR    & IR    & TR    & IR    & TR    & IR    & TR    & IR    & TR    & IR    & TR    & IR \\   \midrule
    C-PGC  & \gc\textbf{90.13} & \gc\textbf{88.82} & \textbf{62.11} & \textbf{64.48} & \textbf{20.53} & \textbf{39.38} & \textbf{43.10} & \textbf{65.93} & \textbf{54.40} & \textbf{72.51} & \textbf{44.79} & \textbf{56.36} \\
    C-PGC$_{CL}$ & \gc76.46    & \gc77.58    & 34.99    & 47.55    & 14.33    & 33.61    & 42.98    & 62.81    & 46.11    & 65.58    & 27.13    & 46.44 \\
    C-PGC$_{Dis}$ & \gc79.54    & \gc82.46    & 56.52    & 62.21    & 20.24    & 38.26    & 39.78    & 65.10     & 52.20     & 71.01    & 42.43    & 55.52 \\
    C-PGC$_{Rand}$ & \gc61.87    & \gc65.17    & 43.69    & 52.54    & 19.51    & 35.47    & 40.33    & 65.77    & 54.15    & 70.62    & 39.43    & 52.59 \\
    C-PGC$_{CA}$ &\gc 85.18    &\gc 83.07    & 45.76    & 53.73    & 15.24    & 34.02    & 39.29    & 60.61    & 47.15    & 40.64    & 32.39    & 48.29 \\ \bottomrule
    \end{tabular} }
\end{table*}%
\begin{figure*}[t]
% \centering
\begin{minipage}[t]{0.35\linewidth}
  \centering
  \includegraphics[width=0.96\textwidth]{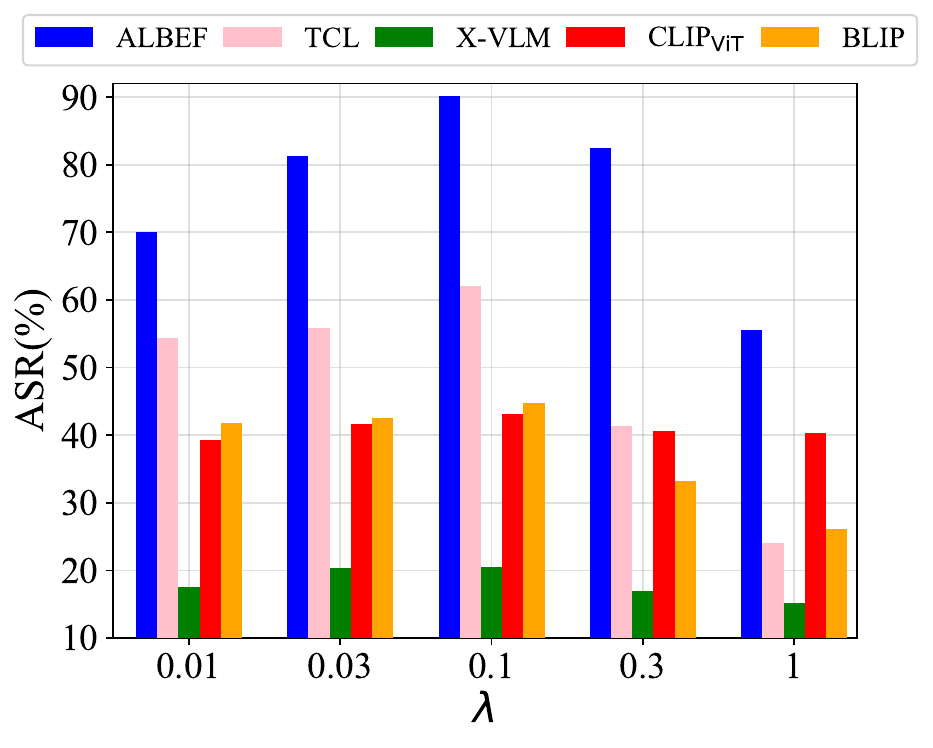}
  \captionof{figure}{ASR of five target models on TR tasks under various values of $\lambda$.} 
  % Note that when $\lambda=0.1$, the ASR of each target model reaches the top value.}
\label{fig:asr_lambda}
\end{minipage}
\begin{minipage}[t]{.015\linewidth}
\quad
\end{minipage}
\begin{minipage}[t]{0.635\linewidth}
  \centering
  \includegraphics[width=\textwidth]{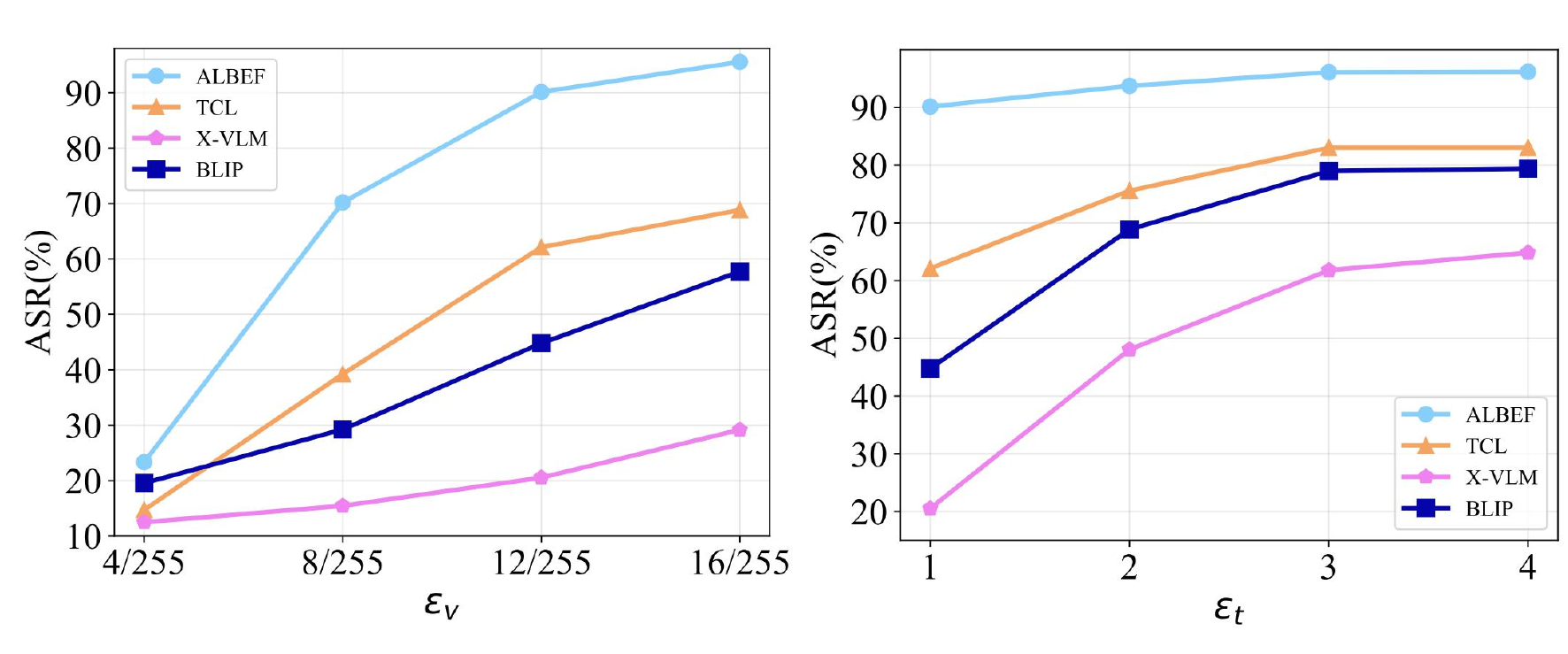}
  \centering
  \captionof{figure}{ASR of five target models on the TR task under different values of perturbation budgets for $\epsilon_{v}$ and $\epsilon_{t}$ respectively.} 
  % As the budget $\epsilon_{v}$ increases, the ASR of each target model is also getting higher.}
\label{fig:eps_v}
\end{minipage}
% \vspace{-1em}
% \vspace{-2em}
 \end{figure*}
% \begin{figure}[t]
% % \centering
%   \centering
%   \includegraphics[width=\linewidth]{imgs/bar_asr_alpha.pdf}
%   \captionof{figure}{ASR of five target models on TR tasks under various $\lambda$.} 
%   % Note that when $\lambda=0.1$, the ASR of each target model reaches the top value.}
% \label{fig:asr_lambda}
% \end{figure}
% \vspace{-1em}
% \vspace{-2em}
As shown in Table \ref{tab:ablation}, the removal of $\mathcal{L}_{CL}$ leads to significant degradation, particularly for black-box transferable attacks. \Eg, a 27.12\% ASR drop occurs in TR tasks when transferring from ALBEF to TCL. This validates the considerable contribution of $\mathcal{L}_{CL}$ to guarantee a successful attack. 
Regarding the influence of $\mathcal{L}_{Dis}$, we demonstrate that the unimodal guidance further enhances the attack on the basis of $\mathcal{L}_{CL}$. Especially in white-box scenarios, $\mathcal{L}_{Dis}$ brings a 10.59\% increase in the ASR of TR tasks on ALBEF. The proposed two loss terms complement each other and jointly underpin the generalizability and transferability of the produced UAP.

% \textbf{The effect of $\mathcal{L}_{Dis}$.} Next, we examine the influence of the unimodal loss $\mathcal{L}_{Dis}$. C-PGC$_{Dis}$ is a version that cancels $\mathcal{L}_{Dis}$ from the training loss. By comparing the ASRs in Table \ref{tab:ablation}, we demonstrate that this term can further enhance the attack capability on the basis of $\mathcal{L}_{CL}$, \eg, a 9.21\% increase in the ASR of TR when transferring from ALBEF to the TCL model, proving the importance of the unimodal guidance. 

\noindent\textbf{The effect of positive sample selection.} 
% To obtain positive samples in contrastive learning, we randomly sample a bunch of data points and select the ones that are farthest from the clean sample. 
To validate the \textit{farthest selection strategy} for positive sample construction, we design another variant C-PGC$_{Rand}$ that adopts randomly sampled data points as positive samples. Results in Table \ref{tab:ablation} reveal the necessity of the proposed selection strategy as it brings an average improvement of 25.96\% in white-box ASR and 4.95\% in black-box ASR.
Moreover, we also find that if the positive samples are not adequately defined, adding the $\mathcal{L}_{CL}$ would even harm the white-box performance (see C-PGC$_{CL}$ and C-PGC$_{Rand}$). This highlights the significance of the proposed selection strategy.
% if the positive samples are not adequately defined, adding $\mathcal{L}_{CL}$ would even cause a negative impact on the white-box attack performance, again verifying the necessity of the proposed farthest samples selection strategy.

% \textbf{The effect of generator $G$.} C-PGC$_{G}$ is the version of C-PGC without the perturbation generator, \ie, we directly randomly initialize and update the adversarial perturbation. The results in Table \ref{tab:ablation} reveal that the absence of the generator's powerful distribution perception capability results in a pronounced reduction in the generalization capability of the learned UAP on test samples, thereby leading to an obvious decrease in the ASR. 
% Specifically, we find that removing the generator will result in a significant performance drop of more than 50\% on average. The generator contributes much more to the final attack performance than all of the remaining loss terms. These findings strongly confirm the crucial importance of introducing the generative network.

\noindent\textbf{The effect of cross-modal conditions.} As aforementioned, cross-attention (CA) modules are introduced into the generator to exploit cross-modal information.
We then design C-PGC$_{CA}$ that cancels these CA layers to explore the influence. As expected, it causes a notable 9.78\% average decrease across six target models, verifying its vital role in boosting attacks. Another finding is that C-PGC$_{CA}$ induces a more pronounced drop in black-box attacks than white-box ones, indicating that cross-modal conditions exert a greater contribution to adversarial transferability.

\noindent\textbf{Different regulatory factor $\lambda$.} The value of $\lambda$ is a critical factor as it adjusts the scales of the two loss terms $\mathcal{L}_{CL}$ and $\mathcal{L}_{Dis}$. We evaluate the attack performance under various values of $\lambda$ to confirm the optimal value. Fig. \ref{fig:asr_lambda} indicates that $\lambda = 0.1$ achieves superior performance.

% \begin{figure}[t]
%   \centering
%   \includegraphics[width=0.95\linewidth]{imgs/bar_asr_alpha.pdf}
%   \centering
%   \caption{ASR of five target models on the TR task under various values of $\alpha$. The surrogate model is ALBEF and the dataset is Flickr30K. Note that when $\alpha=0.1$, the ASR of each target model reaches the top value.}
% \label{fig:asr_alpha}
%  \end{figure}

\noindent\textbf{Different perturbation budgets $\epsilon_{v}$ and $\epsilon_{t}$.} As shown in Fig. \ref{fig:eps_v}, we analyze varying perturbation budgets for $\epsilon_{v}$ and $\epsilon_{t}$. Generally, the ASR increases with the larger perturbation magnitudes. 
Note that when $\epsilon_{v}=4/255$, C-PGC's performance is severely compromised since the budget $4/255$ is too small to allow the UAP to carry enough information required to generalize to diverse data samples. 

It also indicates that the improvement slows down as $\epsilon_v$ increases from $12/255$ to $16/255$. Thus, we select the moderate value of $12/255$ to reach a balance between attack utility and imperceptibility. 
For text perturbation, $\epsilon_t$ exhibits a more profound influence on the black-box attacks. In our experiments, we strictly set $\epsilon_t=1$ for invisibility. Attackers can adjust the value of $\epsilon_t$ in accordance with their demands to trade off the attack efficacy and perturbation stealthiness.

\section{Conclusion}
In this paper, we investigate the challenging task of universal adversarial attacks on VLP models. 
 % and proposes an effective solution that achieves superior performance using only one universal pair of image-text perturbations.
We begin by revealing the deficiency of existing UAP methods and empirically explaining the underlying reasons. 
Based on the analysis, we propose to break the crucial multimodal alignment in VLP models by designing a contrastive-learning generative UAP framework that leverages both unimodal and multimodal information to enhance the attack. 
Extensive experiments validate the efficacy of C-PGC on diverse VLP models and V+L tasks. 
We highlight that the proposed method makes a significant step in exploring classic universal attacks in multimodal attacks and deepens our understanding of the mechanism underlying VLP models. 
We hope our work can promote future research on more sophisticated defenses to further strengthen the robustness of VLP models. 
% against adversarial attacks.
% \newpage
{
    \small
    \bibliographystyle{ieeenat_fullname}
    \bibliography{main}
}
\clearpage
\appendix

\section{Pseudocode of the Proposed Algorithm} \label{appendix:alg}
We present the pseudocode of our proposed attack algorithm for image modality in Alg. \ref{alg:C-PGC}. Note that the text attacks are completely symmetrical as illustrated in Sec. 3.
% \ref{para:textmodal}.

\begin{algorithm}[h]
  \caption{Pseudocode of universal image attacks} \label{alg:C-PGC}
  \begin{algorithmic}[1]
    \Require
      $G_w(\cdot)$: the perturbation generator;
      $D_s$: the multimodal training set;
      $f_I, f_T$: image encoder and text encoder of the surrogate VLP model;
      $K$: the max iteration;
      $\epsilon_v$: the perturbation budget;
      $N$: the scaling times;

    \Ensure
       Universal image perturbation $\delta_v$;
     \State \textbf{Initialize} the fixed noise $z_{v}$ with Gaussian distribution;
     \For{$i \leftarrow 0$ to $K$}
        \State    Randomly sample an image-text pair $(v,\mathbf{t}) \sim D_s$; 
        \State   $\delta_{v} = Clip_{\epsilon_v}(G_{w}(z_v; f_T(\mathbf{t})))$,  $v_{adv} = v + \delta_v$;
        \State   Augment $v$ and $v_{adv}$ into different scales and apply random Gaussian noises to obtain $\mathbf{v}=\{v_1\dots, v_N\}$ and $\mathbf{v}_{adv}=\{v_1^{adv}\dots, v_N^{adv}\}$;
        
        \State   Randomly sample a batch of text sets from $D_s$ and obtain $\mathbf{t}_{pos} = \{t_1'\dots, t_K'\}$ by selecting the one with the farthest feature distance from the clean image $v$; 
        \State Compute $\mathcal{L}_{CL}$ with $\mathbf{v}_{adv}$, $\mathbf{t}$ and $\mathbf{t}_{pos}$ by Eq. (3);
        \State   Compute $\mathcal{L}_{Dis}$ with $\mathbf{v}$ and $\mathbf{v}_{adv}$ by Eq. (4);
        \State Optimize the generator $G_w$ based on Eq. (5);
        \State Backward pass and update $G_w$;
     \EndFor
  \State \textbf{Return} $\delta_v$
  \end{algorithmic}
\end{algorithm}

\section{More Training Details} \label{app:train_detail}
For Flickr30K and MSCOCO, we randomly sample 30,000 images and their captions from the training set to train our perturbation generator. For SNLI-VE and RefCOCO+, we learn the C-PGC directly using their training set with 29,783 and 16,992 images respectively. Since an image corresponds to multiple text descriptions in these datasets, we calculate the average of their textual embedding as the multimodal condition for the cross-attention modules. 
% And we select the text set with the farthest distance to construct the multimodal contrastive loss $\mathcal{L}_{CL}$.

We initialize the noise variable $z_v$ as a $3 \times 3$ matrix. Meanwhile, the initial noise $z_t$'s dimensions in the text modality depend on the size of the hidden layer within the specific VLP model. Concretely, we set its dimension to $1 \times 3$ for ALBEF, TCL, BLIP, and X-VLM, while $1 \times 2$ for the CLIP model. When computing the multimodal contrastive loss $\mathcal{L}_{CL}$, the temperature $\tau$ is set as 0.1. The generator is trained over 40 epochs with the Adam optimizer at a learning rate of $2^{-4}$. 
Following previous works \cite{lu2023set, wang2024transferable}, we employ the attack success rate (ASR) as our quantitative measurement in ITR tasks by computing the extent to which the adversarial perturbations result in victim models' performance deviations from the clean performance.

\section{More Experimental Results} \label{sec:sup_ex}
In this section, we provide more experimental results of our method in various tasks and scenarios.

\begin{table*}[htbp]
  \centering
  \setlength{\tabcolsep}{5.8pt}
  \caption{ASR (\%) of different methods for image-text retrieval tasks on MSCOCO dataset. TR indicates text retrieval based on the input image, while IR is image retrieval using the input text.}
    \resizebox{0.93\linewidth}{!}{\begin{tabular}{cc|cc|cc|cc|cc|cc|cc} \toprule
    \multirow{2}[0]{*}{Source} & \multirow{2}[0]{*}{Method} & \multicolumn{2}{c}{ALBEF} & \multicolumn{2}{c}{TCL} & \multicolumn{2}{c}{X-VLM} & \multicolumn{2}{c}{CLIP$_\textrm{ViT}$} & \multicolumn{2}{c}{CLIP$_\textrm{CNN}$} & \multicolumn{2}{c}{BLIP} \\ \cmidrule(lr){3-14} 
          &        & TR   & IR   & TR   & IR   & TR   & IR   & TR   & IR   & TR   & IR   & TR   & IR \\ \midrule
         \multirow{3}[0]{*}{ALBEF} & GAP      & \gc82.65    & \gc84.35    & 53.6     & 45.46    & 15.09    & 15.64    & 25.18    & 29.94    & 28.06    & 35.28    & 37.44    & 33.61 \\ & ETU      & \gc83.6     & \gc88.98    & 27.43    & 24.47    & 20.39    & 19.94    & 28.54    & 35.05    & 37.01    & 44.72    & 22.25    & 22.03 \\
            & Ours     & \gc\textbf{96.18} & \gc\textbf{95.09} & \textbf{82.49} & \textbf{76.24} & \textbf{39.97} & \textbf{48.58} & \textbf{59.71} & \textbf{67.05} & \textbf{61.27} & \textbf{70.8} & \textbf{59.18} & \textbf{63.89} \\ \midrule
          \multirow{3}[0]{*}{TCL} & GAP      & 55.92    & 48.22    & \gc95.16    & \gc92.29    & 17.34    & 17.01    & 28.73    & 31.19    & 32.27    & 39.81    & 43.59    & 39.64 \\ & ETU      & 60.09    & 50.62    & \gc93.28    & \gc89.28    & 27.19    & 25.41    & 33       & 37.86    & 45.83    & 52.24    & 39.59    & 36.31 \\
                 & Ours     & \textbf{76.62} & \textbf{71.17} & \gc\textbf{96.72} & \gc\textbf{93.88} & \textbf{42.99} & \textbf{48.4} & \textbf{70.32} & \textbf{79.08} & \textbf{74.1} & \textbf{82.97} & \textbf{62.35} & \textbf{66.97}  \\ \midrule
           \multirow{3}[0]{*}{X-VLM} & GAP      & 26.35    & 23.72    & 27.8     & 22.91    & \gc95.1     & \gc88.84    & 32.39    & 38.16    & 52       & 55.4     & 24.67    & 22.65 \\
    & ETU      & 22.94    & 21.63    & 22.01    & 19.65    & \gc96.23    & \gc92.97    & 28.81    & 34.26    & 48.53    & 52.74    & 20.52    & 19.3 \\
    & Ours     & \textbf{51.46} & \textbf{65.71} & \textbf{52.8} & \textbf{64.99} & \gc\textbf{98.89} & \gc\textbf{95.79} & \textbf{67.42} & \textbf{75.45} & \textbf{75.49} & \textbf{82.58} & \textbf{55.74} & \textbf{66.7} \\ \midrule
           \multirow{3}[0]{*}{CLIP$_\textrm{ViT}$} & GAP      & 35.96    & 31.91    & 37.33    & 32.56    & 33.42    & 29.25    & \gc97.71    & \gc96.04    & 74.63    & 74.67    & 33.47    & 31.99 \\
    & ETU      & 31.5     & 29.62    & 33.25    & 30.38    & 32.36    & 29.92    & \gc95.88    & \gc96.34    & \textbf{82.07} & 83.41    & 30.62    & 30.7 \\
    & Ours     & \textbf{46.92} & \textbf{53.89} & \textbf{46.03} & \textbf{50.87} & \textbf{41.49} & \textbf{48.6} & \gc\textbf{98.74} & \gc\textbf{98.01} & 81.58    & \textbf{86.5} & \textbf{47.35} & \textbf{57.55} \\ \midrule
           \multirow{3}[0]{*}{CLIP$_\textrm{CNN}$} & GAP   & 13.57    & 25.21    & 19.05    & 28.87    & 11.59    & 23.13    & 27.46    & 43.16    & \gc73.18    & \gc81.6     & 15.25    & 27.94 \\ & ETU      & 21.71    & 21.92    & 22.33    & 22.8     & 24.77    & 23.93    & 34.99    & 40.3     & \gc\textbf{95.34} & \gc95.14    & 20.06    & 22.26  \\
                 & Ours     & \textbf{33.41} & \textbf{47.96} & \textbf{38.81} & \textbf{50.78} & \textbf{36.59} & \textbf{48.83} & \textbf{66.04} & \textbf{72.59} & \gc94.73    & \gc\textbf{95.21} & \textbf{42.39} & \textbf{57.84} \\ \midrule
           \multirow{3}[0]{*}{BLIP} & GAP    & 12.23    & 23.94    & 14.49    & 25.44    & 6.91     & 17.81    & 20.32    &37.00       & 26.81    & 43.59    & \gc47.21    & \gc73.33 \\ & ETU      & 46.07    & 43.27    & 44.58    & 37.61    & 33.14    & 29.85    & 33.77    & 40.02    & 48.28    & 52.88    & \gc81.27    & \gc83.59 \\
            & Ours     & \textbf{61.95} & \textbf{60.92} & \textbf{60.95} & \textbf{59.57} & \textbf{51.81} & \textbf{52.53} & \textbf{62.23} & \textbf{72.51} & \textbf{69.61} & \textbf{78.44} & \gc\textbf{91.67} & \gc\textbf{90.42}  \\ \bottomrule
    \end{tabular} }
  \label{tab:mscoco}%
\end{table*}%
\begin{table*}[t]
  \centering
  \caption{ASR (\%) of Cross-domain attacks on six models from Flickr30k to MSCOCO and vice versa.}
    \resizebox{0.95\linewidth}{!}{\begin{tabular}{cc|cc|cc|cc|cc|cc|cc} \toprule
    \multirow{2}[0]{*}{Setting} & \multirow{2}[0]{*}{Source} & \multicolumn{2}{c}{ALBEF} & \multicolumn{2}{c}{TCL} & \multicolumn{2}{c}{X-VLM} & \multicolumn{2}{c}{CLIP$_{\textrm{ViT}}$} & \multicolumn{2}{c}{CLIP$_{\textrm{CNN}}$} & \multicolumn{2}{c}{BLIP} \\ \cmidrule(lr){3-14}
             &     & TR   & IR   & TR   & IR   & TR   & IR   & TR   & IR   & TR   & IR   & TR   & IR \\ \midrule
             \multirow{6}[0]{*}{\makecell{Flickr30K\\$\downarrow$\\MSCOCO}} & ALBEF    & \gc\textbf{96.83} & \gc\textbf{94.69} & 81.46    & 74.87    & 44.79    & 51.64    & 63.68    & 73.06    & 69.77    & 78.09    & 68.88    & 70.61 \\
             & TCL      & 78.27    & 73.17    & \gc\textbf{97.83} & \gc\textbf{95.03} & 40.46    & 47.34    & 64.98    & 73.27    & 70.96    & 78.18    & 63.71    & 67.1 \\
             & X-VLM    & 50.63    & 65.91    & 53.23    & 65.65    & \gc\textbf{95.91} & \gc\textbf{93.32} & 65.51    & 74.72    & 75.69    & 81.93    & 57.69    & 67.28 \\
             & CLIP$_\textrm{ViT}$ & 49.88    & 53.39    & 49.47    & 52.21    & 47.77    & 48.52    & \gc\textbf{95.5} & \gc\textbf{97.01} & 83.05    & 85.38    & 50.97    & 57.93 \\
             & CLIP$_\textrm{CNN}$ &  43.05    & 54.19    & 43.04    & 54.39    & 43.73    & 53.94    & 67.3    & 74.39    & \gc\textbf{98.61} & \gc\textbf{97.41} & 47.22    & 59.11 \\
             & BLIP     & 54.45    & 55.51    & 55.63    & 53.02    & 41.07    & 46.93    & 61.69    & 69.24    & 65.52    & 75.23    & \gc\textbf{83.19} & \gc\textbf{82.17} \\ \midrule
             \multirow{6}[0]{*}{\makecell{MSCOCO\\$\downarrow$\\Flickr30K}} & ALBEF    & \gc\textbf{88.08} & \gc\textbf{87.28} & 58.9     & 61.53    & 17.58    & 36.07    & 39.78    & 61.08    & 47.28    & 64.95    & 35.02    & 49.4 \\
             & TCL      & 47.58    & 53.7     & \gc\textbf{87.27} &\gc \textbf{83.55} & 18.6     & 34.45    & 51.85    & 72.22    & 59.46    & 76.09    & 37.75    & 53.08 \\
             & X-VLM    & 25.39    & 46.74    & 27.33    & 49.13    & \gc\textbf{79.98} & \gc\textbf{81.72} & 42.73    & 66.48    & 59.46    & 73.07    & 31.65    & 51.48 \\
             & CLIP$_\textrm{ViT}$ & 21.07    & 39.47    & 24.53    & 42.44    & 15.45    & 36.52    & \gc\textbf{93.97} & \gc\textbf{95.53} & 62.95    & 77.21    & 25.55    & 45.91 \\
             & CLIP$_\textrm{CNN}$ & 13.87    & 37.93    & 19.36    & 41.25    & 15.85    & 37.47    & 42.61    & 66.73    & \gc\textbf{85.75} & \gc\textbf{88.76} & 22.92    & 48.45 \\
             & BLIP     & 33.2     & 46.07    & 36.02    & 47.97    & 23.58    & 38.48    & 43.97    & 65.3     & 56.35    & 71.08    & \gc\textbf{71.91} & \gc\textbf{73.62} \\  \bottomrule
    \end{tabular} } 
  \label{tab:corss_domain_appendix}%
\end{table*}

% \begin{table*}[htbp]
%   \centering
%   \caption{ASR (\%) of C-PGC, C-PGC$_{t}$, and GAP on ITR tasks using Flickr30K. Note that C-PGC$_{t}$ only considers attacking images and thus doesn't apply textual perturbations.}
%     \resizebox{0.85\linewidth}{!}{\begin{tabular}{c|cc|cc|cc|cc|cc|cc}
%     \toprule
%     \multirow{2}[0]{*}{Method} & \multicolumn{2}{c}{ALBEF} & \multicolumn{2}{c}{TCL} & \multicolumn{2}{c}{X-VLM} & \multicolumn{2}{c}{CLIP$_\textrm{ViT}$} & \multicolumn{2}{c}{CLIP$_\textrm{CNN}$} & \multicolumn{2}{c}{BLIP} \\  \cmidrule(lr){2-13}
%              & TR       & IR       & TR       & IR       & TR       & IR       & TR       & IR       & TR       & IR       & TR       & IR \\ \midrule
%     GAP      & \gc69.78    & \gc81.59    & 22.15    & 29.97    & 6.61     & 18.37    & 23.4     & 37.54    & 29.92    & 44.29    & 16.09    & 28.12 \\ 

%     ETU      & \gc78.01    & \gc84.56    & 29.92    & 35.91    & 14.33     & 22.03    & 23.77     & 39.2    & 33.55    & 47.69    & 22.61    & 32.28\\
%     C-PGC$_{t}$   & \gc86.74    & \gc86.3     & 50.1     & 50.2     & 10.87    & 21.53    & 26.28    & 39.3     & 33.42    & 48.32    & 31.55    & 36.77 \\ 
%     C-PGC    & \gc\textbf{90.13} & \gc\textbf{88.82} & \textbf{62.11} & \textbf{64.48} & \textbf{20.53} & \textbf{39.38} & \textbf{43.1} & \textbf{65.93} & \textbf{54.4} & \textbf{72.51} & \textbf{44.79} & \textbf{56.36} \\ \bottomrule
%     \end{tabular}
%     }
%   \label{tab:wotext}
% \end{table*}

\begin{figure}[htbp]
  \centering
  % \caption{Performance of image captioning. The baseline represents the original performance of the target model on clean data and the target model is BLIP.}
  \includegraphics[width=0.7\linewidth]{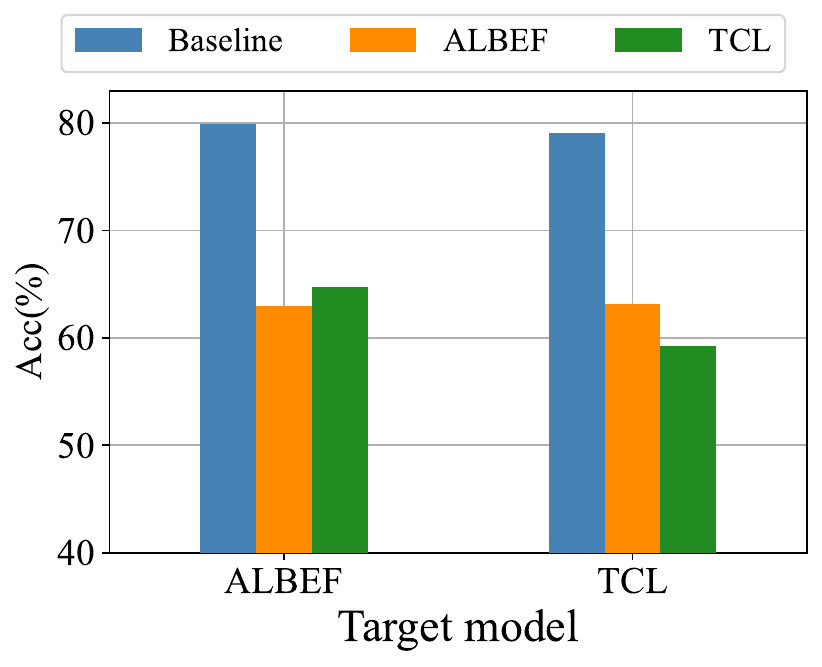}
  \captionof{figure}{Accuracy of VE tasks for different source and target models. The Baseline indicates the clean performance.}
\label{fig:ve}
\end{figure}

\textbf{Visual entailment tasks.}  \label{sec:VE} Given an image and a textual description, visual entailment involves determining whether the textual description can be inferred from the semantic information of the image. We align with previous VLP attacks \citep{zhang2022towards, wang2024transferable} and conduct experiments on the SNLI-VE \citep{xie2019visual} dataset using the ALBEF and TCL models. Note that the Baseline represents the clean performance of the target model on the clean data and the orange and green indicate ALBEF and TCL as source models respectively. 
The results presented in Fig. \ref{fig:ve} reveal that C-PGC obtains impressive attack effects by decreasing the average accuracy by nearly 20\%. 

Notably, \citep{do2020snli} has reported a large number of annotation errors in the labels of the SNLI-VE corpus used for VE tasks. Therefore, the presented results are only for experimental integrity and reference purposes.

\textbf{Image-text retrievals on MSCOCO dataset.} We then supplement the ASR of the ITR tasks on the MSCOCO dataset in Table \ref{tab:mscoco}. 
The results again reveal that C-PGC greatly enhances the attack. Particularly in the more realistic and challenging transferable scenarios, the proposed method achieves considerably better performance, \eg, 82.49\% and 76.24\% increase in ASR of TR and IR tasks when transferring from ALBEF to TCL, confirming the superiority of our contrastive learning-based paradigm. 

\textbf{Cross-domain scenarios.} We proceed to discuss the attack performance of the proposed algorithm in a more challenging scenario where there is an obvious distribution shift between the training dataset and the test samples. Specifically, we generate universal adversarial perturbations based on MSCOCO or Flickr30K and evaluate them accordingly on the other dataset. We present the attack success rates on the retrieval tasks across six models in Table \ref{tab:corss_domain_appendix}. It can be observed that the domain gap indeed has a negative effect on attack performance. However, our method still maintains excellent ASR in most cases, unveiling its outstanding cross-domain transferability. 

% \textbf{Ablation study of the text perturbation.} We introduce another variant C-PGC$_{t}$ that cancels the perturbation from the text side to investigate the contribution of image perturbation, text perturbation, and their synergy.
% The comparison results of C-PGC, C-PGC$_{t}$, and GAP using ALBEF as the surrogate model are shown in Table \ref{tab:wotext}. 

% We find that when merely applying image perturbations  (C-PGC$_{t}$), our design still outperforms GAP with notable improvements, validating the proposed techniques' effectiveness in enhancing the image perturbation.
% Moreover, the superiority of C-PGC over C-PGC$_{t}$ indicates that the incorporation of textual perturbations can further boost the universal attacks on the basis of C-PGC$_{t}$ since the text perturbation facilitates the deconstruction of the learned cross-modal alignment.
\textbf{Results of R@5 and R@10.} \label{sec:r5_r10}
As aforementioned, we supplement the ASR of the ITR tasks based on R@5 and R@10 metrics and provide the attack success rates in Table \ref{tab:R5R10}. 
Obviously, our proposed C-PGC still consistently attains better performance than the baseline method ETU, regardless of the evaluation measurements for retrieval results.

\begin{table*}[htbp]
  \centering
  \caption{Attack success rates (\%) regarding R@5 and R@10 metrics of our C-PGC and ETU for image-text retrieval tasks.}
    \resizebox{\linewidth}{!}{\begin{tabular}{ccc|cc|cc|cc|cc|cc|cc} \toprule
    \multirow{2}[0]{*}{Dataset} & \multirow{2}[0]{*}{Source} & \multirow{2}[0]{*}{Method} & \multicolumn{2}{c}{ALBEF} & \multicolumn{2}{c}{TCL} & \multicolumn{2}{c}{X-VLM} & \multicolumn{2}{c}{CLIP$_\textrm{ViT}$} & \multicolumn{2}{c}{CLIP$_\textrm{CNN}$} & \multicolumn{2}{c}{BLIP} \\ \cmidrule(lr){4-15} 
             &       &       & TR   & IR   & TR   & IR   & TR   & IR   & TR   & IR   & TR   & IR   & TR   & IR \\ \midrule
    \multirow{12}[0]{*}{\makecell{Flickr30K\\(R@5)}} & \multirow{2}[0]{*}{ALBEF} & ETU      & \gc68.54    & \gc77.68    & 14.41    & 14.43    & 4.1      & 6.41     & 6.11     & 16.97    & 11.57    & 23.11    & 9.36     & 13.3 \\
             &          & Ours     & \gc\textbf{83.67} & \gc\textbf{80.02} & \textbf{41.84} & \textbf{42.18} & \textbf{6.9} & \textbf{17.19} & \textbf{18.34} & \textbf{41.03} & \textbf{26.22} & \textbf{49.42} & \textbf{24.25} & \textbf{34.59} \\ \cmidrule(lr){2-15}
             & \multirow{2}[0]{*}{TCL} & ETU      & 14.43    & 16.23    & \gc84.88    & \gc80.43    & 2.4      & 5.8      & 5.6      & 17.37    & 13.27    & 24.27    & 7.24     & 10.98 \\
             &          & Ours     & \textbf{29.76} & \textbf{35.62} & \gc\textbf{90.89} & \gc\textbf{84.18} & \textbf{3.2} & \textbf{13.65} & \textbf{20.93} & \textbf{42.06} & \textbf{25.27} & \textbf{49.1} & \textbf{16.5} & \textbf{30.32} \\ \cmidrule(lr){2-15}
             & \multirow{2}[0]{*}{X-VLM} & ETU      & 3.81     & 5.85     & 3.9      & 6.51     & \gc89.2     & \gc84.57    & 5.18     & 16.9     & 14.33    & 24.52    & 3.12     & 7.81 \\
             &          & Ours     & \textbf{7.62} & \textbf{25.1} & \textbf{8.71} & \textbf{26.63} & \gc\textbf{89.2} & \gc\textbf{85.84} & \textbf{19.38} & \textbf{42.48} & \textbf{30.89} & \textbf{50.7} & \textbf{13.68} & \textbf{29} \\ \cmidrule(lr){2-15}
             & \multirow{2}[0]{*}{CLIP$_\textrm{ViT}$} & ETU & 5.31     & 7.88     & 7.51     & 10.42    & 4.7      & 8.92     & \gc76.42    & \gc84.54    & 35.46    & 41.73    & 5.03     & 10.2 \\
             &          & Ours     & \textbf{6.31} & \textbf{17.51} & \textbf{8.01} & \textbf{19.65} & \textbf{4.3} & \textbf{15.1} & \gc\textbf{76.89} & \gc\textbf{85.2} & \textbf{39.6} & \textbf{54.68} & \textbf{9.15} & \textbf{23.23} \\ \cmidrule(lr){2-15}
             & \multirow{2}[0]{*}{CLIP$_\textrm{CNN}$} & ETU      & 1.7      & 5.02     & 3.4      & 7.16     & 1.4      & 6.02     & 6.22     & 17.61    & \gc84.39    & \gc87.85    & 2.31     & 8.26 \\
             &          & Ours     & \textbf{4.41} & \textbf{19.8} & 6.41     & \textbf{25.19} & \textbf{4.8} & \textbf{23.49} & \textbf{18.76} & \textbf{43.82} & \gc\textbf{90.34} & \gc\textbf{88.12} & \textbf{8.95} & \textbf{26.89} \\ \cmidrule(lr){2-15}
             & \multirow{2}[0]{*}{BLIP} & ETU      & 8.22     & 10.11    & 6.71     & 10.46    & 3.2      & 5.62     & 5.28     & 16.39    & 9.24     & 21.97    & \gc45.98    & \gc73.75 \\
             &          & Ours     & \textbf{14.43} & \textbf{21.67} & \textbf{13.91} & \textbf{21.59} & \textbf{5.4} & \textbf{14.54} & \textbf{18.03} & \textbf{36.26} & \textbf{23.89} & \textbf{44.79} & \gc\textbf{59.26} & \gc\textbf{74.82} \\ \midrule
    \multirow{12}[0]{*}{\makecell{MSCOCO\\(R@5)}} & \multirow{2}[0]{*}{ALBEF} & ETU      & \gc81.73    & \gc88.76    & 13.45    & 11.51    & 9.01     & 9.32     & 16.85    & 20.74    & 22.6     & 28.86    & 10.96    & 12.02 \\
             &          & Ours     & \gc\textbf{93.36} & \gc\textbf{91.56} & \textbf{70.76} & \textbf{62.31} & \textbf{19.97} & \textbf{30.46} & \textbf{41.58} & \textbf{51.23} & \textbf{44.14} & \textbf{55.98} & \textbf{41.08} & \textbf{49.22} \\ \cmidrule(lr){2-15}
             & \multirow{2}[0]{*}{TCL} & ETU      & 44.94    & 36.45    & \gc90.21    & \gc84.54    & 13.46    & 13.1     & 19.52    & 23.03    & 30.81    & 36.37    & 25.63    & 25.11  \\
             &          & Ours     & \textbf{60.62} & \textbf{56.21} & \gc\textbf{94.89} & \gc\textbf{90.33} & \textbf{22.08} & \textbf{30.38} & \textbf{53.14} & \textbf{64.98} & \textbf{58.85} & \textbf{70.77} & \textbf{45.28} & \textbf{53.55} \\ \cmidrule(lr){2-15}
             & \multirow{2}[0]{*}{X-VLM} & ETU      & 11.03    & 11.11    & 10.22    & 9.24     & \gc94.36    & \gc90.02    & 17.56    & 20.8     & 33.45    & 38.14    & 10.08    & 10.12 \\
             &          & Ours     & \textbf{31.59} & \textbf{48.69} & \textbf{32.1} & \textbf{48.11} & \gc\textbf{96.7} & \gc\textbf{91.66} & \textbf{49.53} & \textbf{60.82} & \textbf{59.83} & \textbf{69.59} & \textbf{37.4} & \textbf{52.5} \\ \cmidrule(lr){2-15}
             & \multirow{2}[0]{*}{CLIP$_\textrm{ViT}$} & ETU      & 15.89    & 16.01    & 18.61    & 16.11    & 16.45    & 16.61    & \gc93.12    & \gc94.66    & \textbf{72.97}    & 75.5     & 17.1     & 19.26 \\
             &          & Ours     & \textbf{25.69} & \textbf{35.95} & \textbf{24.69} & \textbf{33.14} & \textbf{21.37} & \textbf{31.38} & \gc\textbf{96.7} & \gc\textbf{96.49} & 70.76 & \textbf{77.86} & \textbf{28.72} & \textbf{42.01} \\ \cmidrule(lr){2-15}
             & \multirow{2}[0]{*}{CLIP$_\textrm{CNN}$} & ETU      & 9.55     & 10.34    & 9.9      & 10.95    & 10.98    & 11.96    & 22.63    & 26.84    & \gc90.7     & \gc\textbf{92.07} & 9.69     & 12.25 \\
             &          & Ours     & \textbf{16.83} & \textbf{31} & \textbf{19.86} & \textbf{34.54} & \textbf{18.84} & \textbf{34.02} & \textbf{50.21} & \textbf{59.2} & \gc\textbf{90.94} & \gc90.43    & \textbf{25.5} & \textbf{44.89} \\ \cmidrule(lr){2-15}
             & \multirow{2}[0]{*}{BLIP} & ETU      & 29.6     & 29.6     & 26.95    & 21.97    & 17.67    & 16.14    & 20.83    & 24.27    & 32.77    & 36.72    & \gc76.16    & \gc80.69 \\
             &          & Ours     & \textbf{42.56} & \textbf{43.73} & \textbf{41.72} & \textbf{41.8} & \textbf{31.05} & \textbf{35.63} & \textbf{44.37} & \textbf{57.9} & \textbf{54.47} & \textbf{66.01} & \gc\textbf{81.71} & \gc\textbf{81.91} \\ \midrule
    \multirow{12}[0]{*}{\makecell{Flickr30K\\(R@10)}} & \multirow{2}[0]{*}{ALBEF} & ETU      & \gc65.8     & \gc74.89    & 10       & 9.36     & 2.8      & 3.79     & 2.43     & 11.04    & 6.95     & 15.79    & 6.62     & 8.57 \\
             &          & Ours     & \gc\textbf{80.5} & \gc\textbf{75.17} & \textbf{34.8} & \textbf{34.28} & \textbf{4.2} & \textbf{11.72} & \textbf{9.83} & \textbf{31.14} & \textbf{16.87} & \textbf{39.40 } & \textbf{18.76} & \textbf{27.08} \\ \cmidrule(lr){2-15}
             & \multirow{2}[0]{*}{TCL} & ETU      & 11.4     & 11.35    & \gc82.2     & \gc77.33    & 1.6      & 3.38     & 3.04     & 10.63    & 7.06     & 17.14    & 4.91     & 7.34 \\
             &          & Ours     & \textbf{24.2} & \textbf{27.32} & \gc\textbf{89.2} & \gc\textbf{80.73} & \textbf{2.1} & \textbf{9.33} & \textbf{12.77} & \textbf{32.4} & \textbf{16.97} & \textbf{38.63} & \textbf{12.54} & \textbf{24.1} \\ \cmidrule(lr){2-15}
             & \multirow{2}[0]{*}{X-VLM} & ETU      & 1.9      & 3.48     & 2.4      & 3.48     & \gc\textbf{87.1} & \gc81.87    & 3.14     & 10.91    & 8.59     & 16.66    & 2.01     & 4.72 \\
             &          & Ours     & \textbf{4.1} & \textbf{17.79} & \textbf{4.6} & \textbf{19.27} & \gc\textbf{86.3} & \gc\textbf{82.94} & \textbf{11.14} & \textbf{31.49} & \textbf{21.06} & \textbf{40.3} & \textbf{7.32} & \textbf{21.81} \\ \cmidrule(lr){2-15}
             & \multirow{2}[0]{*}{CLIP$_\textrm{ViT}$} & ETU      & 3.6      & 4.75     & 4.6      & 6.43     & 2.7      & 5.47     & \gc\textbf{68.17} & \gc78.93    & 27.2     & 32.94    & 3.11     & 6.35 \\
             &          & Ours     & \textbf{4.2} & \textbf{11.45} & \textbf{4.6} & \textbf{13.08} & \textbf{2.8} & \textbf{10.15} & \gc67.98 & \gc\textbf{79.46} & \textbf{29.75} & \textbf{45.56} & \textbf{5.52} & \textbf{17.23} \\ \cmidrule(lr){2-15}
             & \multirow{2}[0]{*}{CLIP$_\textrm{CNN}$} & ETU      & 0.6      & 2.64     & 1.6      & 3.86     & 0.7      & 3.48     & 3.65     & 11.32    & \gc80.37    & \gc\textbf{85.18} & 1.2      & 5.31 \\
             &          & Ours     & \textbf{2.4} & \textbf{14.59} & \textbf{3.5} & \textbf{18.36} & \textbf{2.3} & \textbf{17.95} & \textbf{11.75} & \textbf{34.23} & \gc\textbf{86.4} & \gc83.83    & \textbf{5.52} & \textbf{20.86} \\ \cmidrule(lr){2-15}
             & \multirow{2}[0]{*}{BLIP} & ETU      & 6.5      & 6.52     & 4.9      & 6.57     & 1.8      & 3.32     & 2.63     & 10.52    & 5.52     & 14.57    & \gc42.43    & \gc71.26 \\
             &          & Ours     & \textbf{11.2} & \textbf{15.49} & \textbf{9.1} & \textbf{14.14} & \textbf{2.8} & \textbf{10.19} & \textbf{9.83} & \textbf{27.46} & \textbf{14.83} & \textbf{34.43} & \gc\textbf{53.46} & \gc\textbf{72} \\ \midrule
    \multirow{12}[0]{*}{\makecell{MSCOCO\\(R@10)}} & \multirow{2}[0]{*}{ALBEF} & ETU      & \gc81.14    & \gc87.58    & 9.08     & 7.92     & 5.44     & 6.33     & 12.62    & 16.15    & 17.71    & 23.05    & 7.63     & 9.15 \\
             &          & Ours     & \gc\textbf{91.58} & \gc\textbf{89.62} & \textbf{64.5} & \textbf{55.3} & \textbf{13.81} & \textbf{23.34} & \textbf{33.3} & \textbf{43.75} & \textbf{35.82} & \textbf{48.38} & \textbf{33.77} & \textbf{43.11} \\ \cmidrule(lr){2-15}
             & \multirow{2}[0]{*}{TCL} & ETU      & 38.6     & 30.39    & \gc88.56    & \gc82.6     & 8.92     & 9.3      & 14.61    & 18.43    & 25.24    & 30.06    & 20.37    & 21.41 \\
             &          & Ours     & \textbf{52.59} & \textbf{49.09} & \gc\textbf{93.63} & \gc\textbf{88.53} & \textbf{15.04} & \textbf{23.25} & \textbf{44.22} & \textbf{58.02} & \textbf{50.16} & \textbf{63.95} & \textbf{37.77} & \textbf{47.26} \\ \cmidrule(lr){2-15}
             & \multirow{2}[0]{*}{X-VLM} & ETU      & 7.41     & 7.37     & 6.76     & 6.27     & \gc93.17    & \gc88.66    & 12.93    & 16.43    & 28.32    & 32.47    & 6.73     & 8.01 \\
             &          & Ours     & \textbf{23.01} & \textbf{40.39} & \textbf{23.15} & \textbf{40.07} & \gc\textbf{94.97} & \gc\textbf{88.95} & \textbf{40.24} & \textbf{53.74} & \textbf{52} & \textbf{62.7} & \textbf{30.43} & \textbf{45.67} \\ \cmidrule(lr){2-15}
             & \multirow{2}[0]{*}{CLIP$_\textrm{ViT}$} & ETU      & 11.05    & 11.5     & 13.67    & 11.67    & 11.21    & 12.14    & \gc91.47    & \gc93.8     & \textbf{67.7}     & 71.14    & 13.03    & 15.4 \\
             &          & Ours     & \textbf{17.87} & \textbf{28.52} & \textbf{17.48} & \textbf{26.09} & \textbf{14} & \textbf{24.67} & \gc\textbf{95.55} & \gc\textbf{95.31} & 64.04 & \textbf{72.75} & \textbf{22.05} & \textbf{35.65} \\ \cmidrule(lr){2-15}
             & \multirow{2}[0]{*}{CLIP$_\textrm{CNN}$} & ETU      & 6.05     & 7.02     & 6.54     & 7.54     & 6.97     & 8.36     & 17.65    & 21.66    & \gc88.13    & \gc\textbf{90.16} & 6.73     & 9.45 \\
             &          & Ours     & \textbf{10.77} & \textbf{24.11} & \textbf{13.34} & \textbf{27.53} & \textbf{12.31} & \textbf{28.03} & \textbf{41.33} & \textbf{52.02} & \gc\textbf{88.28} & \gc87.14    & \textbf{20.26} & \textbf{39.43} \\ \cmidrule(lr){2-15}
             & \multirow{2}[0]{*}{BLIP} & ETU      & 23.63    & 24.4     & 19.69    & 16.37    & 11.55    & 11.86    & 16.3     & 19.66    & 26.04    & 30.64    & \gc73.88    & \gc77.63 \\
             &          & Ours     & \textbf{33.64} & \textbf{36.14} & \textbf{32.15} & \textbf{33.8} & \textbf{22.64} & \textbf{28.52} & \textbf{36.07} & \textbf{50.3} & \textbf{47} & \textbf{59.07} & \gc\textbf{78.39} & \gc\textbf{78.98} \\  \bottomrule
    \end{tabular} }
  \label{tab:R5R10}%
\end{table*}%

\begin{table*}[t]
  \centering
  \caption{ASR (\%) of ITR tasks under defense strategies. The surrogate model is ALBEF and the dataset is Flick30K. LT denotes the LanguageTool that corrects adversarial words within the sentence.}
    \resizebox{0.88\linewidth}{!}{\begin{tabular}{c|cc|cc|cc|cc|cc|cc} \toprule
    \multirow{2}[0]{*}{Method} & \multicolumn{2}{c}{ALBEF} & \multicolumn{2}{c}{TCL} & \multicolumn{2}{c}{X-VLM} & \multicolumn{2}{c}{CLIP$_\textrm{ViT}$} & \multicolumn{2}{c}{CLIP$_\textrm{CNN}$} & \multicolumn{2}{c}{BLIP} \\ \cmidrule(lr){2-13}
          & TR   & IR   & TR   & IR   & TR   & IR   & TR   & IR   & TR   & IR   & TR   & IR \\ \midrule
    Gaussian & 37.92    & 49.49    & 32.4     & 47.04    & 19.31    & 37.79    & 42.49    & 65.61    & 50       & \textbf{72.23}    & 29.65    & 48.77 \\
    Medium & 53.13    & 61.6     & 39.54    & 51.96    & 20.43    & 39.69    & 46.31    & 66.92    & 57.9     & 74.51    & 33.75    & 52.68 \\
    Average & 29.09    & 44.91    & 29.61    & 44.72    & 17.89    & 36.07    & 42.98    & \textbf{65.42}    & \textbf{49.74}    & 72.48    & \textbf{27.55}    & \textbf{46.9} \\
    JPEG  & 59.3     & 63.7     & 42.34    & 52.52    & 21.65    & 41.58    & \textbf{41.26}    & 65.77    & 53.5     & 72.62    & 37.01    & 55.04 \\
    DiffPure & 64.34    & 74.63    & 65.22    & 74.8     & 66.06    & 75.19    & 78.08    & 86.7     & 82.25    & 88.03    & 70.45    & 79.09 \\
    NRP   & 32.33    & 40.63    & \textbf{20.19}    & 39.23    & \textbf{14.63}    & 32.62    & 48.4     & 69       & 59.72    & 74.09    & 30.28    & 52.2 \\
    NRP+LT & \textbf{29.05}    & \textbf{35.23}    & 21.33    & \textbf{37.41}    & 15.55    & \textbf{29.63}    & 47.19    & 67.35    & 56.82    & 73.47    & 28.23    & 50.59 \\
 \bottomrule
    \end{tabular} }
  \label{tab:defense}%
\end{table*}%
\begin{table*}[t]
  \centering
  \caption{ASR results of the proposed method with different loss functions on Flickr30 when the surrogate model is ALBEF.}
    \resizebox{0.92\linewidth}{!}{\begin{tabular}{ccccccccccccc} \toprule
    \multirow{2}[0]{*}{Method} & \multicolumn{2}{c}{ALBEF} & \multicolumn{2}{c}{TCL} & \multicolumn{2}{c}{X-VLM} & \multicolumn{2}{c}{CLIP$_\textrm{ViT}$} & \multicolumn{2}{c}{CLIP$_\textrm{CNN}$} & \multicolumn{2}{c}{BLIP} \\ \cmidrule(lr){2-13}
          & TR    & IR    & TR    & IR    & TR    & IR    & TR    & IR    & TR    & IR    & TR    & IR \\ \midrule
    $\mathcal{L}_{MSE}$   & \gc12.02 & \gc30.75 & 14.39 & 35.08 & 11.41 & 30.79 & 37.32 & 56.05 & 40.17 & 56.39 & 19.66 & 37.33 \\
    $\mathcal{L}_{Cos}$   & \gc57.55 & \gc67.4  & 37.06 & 49.45 & 10.7  & 28.48 & 37.49 & 58.3  & 40.87 & 58.39 & 23.33 & 39.44 \\
    $\mathcal{L}_{CL}$    & \gc\textbf{76.46} & \gc\textbf{82.46} & \textbf{56.52} & \textbf{62.61} & \textbf{14.33} & \textbf{33.61} & \textbf{42.98} & \textbf{62.81} & \textbf{46.11} & \textbf{65.58} & \textbf{27.13} & \textbf{46.44} \\ \midrule
    $\mathcal{L}_{MSE}$+$\mathcal{L}_{Dis}$ & \gc81.09 & \gc83.71 & 48.76 & 56.54 & 17.58 & 35.72 & 41.5  & 64.72 & 47.41 & 70.34 & 35.96 & 51.76 \\
    $\mathcal{L}_{Cos}$+$\mathcal{L}_{Dis}$ & \gc65.20  & \gc72.71 & 36.13 & 50.06 & 18.63 & 36.74 & 42.23 & 65.17 & 50.91 & 69.78 & 36.91 & 50.69 \\
    $\mathcal{L}_{CL}$+$\mathcal{L}_{Dis}$  & \gc\textbf{90.13} & \gc\textbf{88.82} & \textbf{62.11} & \textbf{64.48} & \textbf{20.53} & \textbf{39.38} & \textbf{43.1} & \textbf{65.93} & \textbf{54.4} & \textbf{72.51} & \textbf{44.79} & \textbf{56.36} \\ \bottomrule
    \end{tabular}}
  \label{tab:loss_f}%
\end{table*}%

\textbf{Performance under Defenses.} 
We next analyze several defense strategies to mitigate the potential harm from C-PGC.
Concretely, we totally align with TMM \citep{wang2024transferable} and consider several input preprocessing-based schemes, including image smoothing \citep{ding2019advertorch} (Gaussian, medium, average smoothing), JPEG compression \citep{dziugaite2016study}, NRP \citep{naseer2020self}, and the DiffPure \citep{nie2022diffusion}, a powerful purification defense using diffusion models. For adversarial text correction, we choose the LanguageTool (LT) \citep{wang2024transferable}, which has been widely adopted in various scenarios due to its universality and effectiveness. 

The attack results in Table \ref{tab:defense} demonstrate that the proposed attack still attains great ASR against different powerful defenses. It also indicates that NRP+LT would be a decent choice to alleviate the threat brought by C-PGC. Another noteworthy finding is that, although DiffPure \citep{nie2022diffusion} exhibits remarkable performance in defending attacks in classification tasks, its ability is greatly reduced in V+L scenarios since the denoising process could also diminish some texture or semantic information that is critical for VLP models, thereby acquiring unsatisfactory defense effects.

\section{Rationality behind the Loss Design}
\label{sec:rationale}
{It is widely acknowledged that contrastive learning serves as a powerful and foundational tool for modality alignment in VLP models, establishing a nearly point-to-point relationship between image and text features. 
% Our core idea stems from the general principle: \textit{``It's easier to tear down than to build up."} 
Since contrastive learning can establish robust and precise alignment, leveraging the same technique to disrupt the established alignments is also promising to yield effective performance.}

Taking image attack as an example, the underlying principle behind our contrastive learning-based attack can be understood from two perspectives:
\begin{itemize}[leftmargin=*]
\item Leverage the originally matched texts as negative samples to push aligned image-text pairs apart. This broadly corresponds to the objective of traditional untargeted adversarial attacks. 
\item Additionally, the proposed paradigm introduces dissimilar texts as positive samples to further pull the adversarial image 
 out of its original subspace and relocate it to an incorrect feature area.
 \end{itemize}
 By simultaneously harnessing the collaborative effects of \textit{push} (negative samples) and \textit{pull} (positive samples), the proposed contrastive framework achieves exceptional attack performance, which has been validated by comprehensive experimental results.
Besides, we also explore several potential alternative loss functions that more directly align with the common goal of untargeted attack in Table \ref{tab:loss_f}, including maximizing the cosine distance $\mathcal{L}_{Cos}$ or MSE distance $\mathcal{L}_{MSE}$ between features of matched image-text pairs.

{Recall that $\mathcal{L}_{CL}$ and $\mathcal{L}_{Dis}$ denote our designed contrastive loss and unimodal loss terms respectively. As observed, the integration of $\mathcal{L}_{CL}$ consistently brings significant ASR improvements, verifying the rationality and superiority of the adopted contrastive loss.}

% \section{Comparison with a Concurrent Study}
% We notice a concurrent study \citep{zhang2024universal} on UAP attacks for VLP models, which also shows promising attack performance. To make a fair comparison, we faithfully reproduce this algorithm using their publicly released code under the same experimental settings as ours. Note that \cite{zhang2024universal} implements several versions of their method and we report their best results in Table \ref{tab:etu}.

% {By contrastively training the conditional generator, the proposed C-PGC greatly enhances the attack by achieving significant improvements in ASR. Particularly in the more realistic and challenging transferable scenarios, the proposed method achieves considerably better performance, e.g., 32.19\% and 28.57\% increase in ASR of TR and IR tasks when transferring from ALBEF to TCL. These results strongly confirm the superiority of our contrastive learning-based generative paradigm.}
\begin{table*}[htbp]
  \centering
  \caption{Comparison of BERTScore between clean and adversarial texts across different surrogate models.}
    \resizebox{0.9\linewidth}{!}{\begin{tabular}{c|ccc|ccc|ccc|ccc} \toprule
    \multirow{2}[0]{*}{Method} & \multicolumn{3}{c}{ALBEF} & \multicolumn{3}{c}{TCL} & \multicolumn{3}{c}{CLIP$_\textrm{ViT}$} & \multicolumn{3}{c}{CLIP$_\textrm{CNN}$} \\ \cmidrule(lr){2-13}
          & P     & R     & F1    & P     & R     & F1    & P     & R     & F1    & P     & R     & F1 \\ \midrule
    Co-Attack \cite{zhang2022towards} & 0.8328 & 0.8589 & 0.8455 & 0.8325 & 0.8588 & 0.8453 & 0.8269 & 0.8526 & 0.8394 & 0.8271 & 0.8530 & 0.8397 \\
    SGA \cite{lu2023set}  & 0.8389 & \textbf{0.8654} & 0.8518 & 0.8376 & 0.8646 & 0.8509 & 0.8416 & \textbf{0.8697} & 0.8553 & 0.8378 & 0.8650 & 0.8511 \\
    Ours  & \textbf{0.8891} & 0.8613 & \textbf{0.8748} & \textbf{0.8924} & \textbf{0.8687} & \textbf{0.8802} & \textbf{0.8746} & 0.8684 & \textbf{0.8713} & \textbf{0.8948} & \textbf{0.8842} & \textbf{0.8893} \\ \bottomrule
    \end{tabular}}
  \label{tab:BertScore}
\end{table*}
\section{Semantic Similarity Analysis}
The basic objective of untargeted adversarial attacks is to fool the victim model to output incorrect predictions \citep{dong2018boosting}, while the attacker is supposed to preserve semantic similarity between original and adversarial samples to ensure attack imperceptibility. 
In our implementation, we follow the rigorous setup in prior works \citep{zhang2022towards,lu2023set,wang2024transferable} that modify only one single word to preserve the attack stealthiness. To quantitatively analyze the influence, we provide the BERT scores \citep{zhangbertscore}, which calculate the P (precision), R (recall), and F1 (F1 score), to measure the semantic distance between 5,000 clean and adversarial sentences in Table \ref{tab:BertScore}. Note that we provide existing well-acknowledged sample-specific algorithms Co-Attack \cite{zhang2022towards} and SGA \citep{lu2023set} as references.
% Table generated by Excel2LaTeX from sheet 'rebuttal'
 
As observed, C-PGC generally acquires higher similarity scores than existing sample-specific methods across various surrogate models, which validates that C-PGC achieves an eligible perturbation strategy in terms of text perturbation imperceptibility. 
Basically, the lower semantic similarity of sample-specific approaches stems from their word-selection mechanism, which maximizes the semantic distance tailored to every input sentence for attack enhancement. \Ie, these methods select the adversarial word that maximizes the distance between the original and perturbed texts for every input sentence, which inherently leads to relatively larger semantic deviations. This highlights that our method achieves a better balance between efficacy and stealthiness.
% \begin{table}[t]
%   \centering
%   \caption{Comparison of BLEU metrics between clean and adversarial texts.}
%     \resizebox{\linewidth}{!}{\begin{tabular}{cccccc} \toprule
%     Method & B@4 & METEOR & ROUBE\_L & CIDEr & SPICE \\ \midrule
%     Co-Attack & 0.79  & 0.52  & 0.895 & 7.03  & 0.661 \\
%     SGA   & 0.798 & 0.527 & 0.898 & 7.159 & 0.668 \\
%     Ours  & \textbf{0.889} & \textbf{0.552} & \textbf{0.905} & \textbf{8.036} & \textbf{0.671} \\ \bottomrule
%     \end{tabular}}
%   \label{tab:BLEU}%
% \end{table}%
% Besides, we also provide BLEU metrics when the surrogate model is ALBEF in Table \ref{tab:BLEU}. These results again validate the better stealthiness of our C-PGC.

\section{Multimodal Alignment Destruction}
 To provide more intuitive evidence that our C-PGC successfully destroys the image-text alignment relationship, we compute the distance between the encoded image and text embeddings before and after applying the UAP. For an input pair (v, t), we calculate the relative distance $d_{rel}$ by:
\begin{equation}
d_{rel}=\frac{||(f_I(v + \delta_v)- f_T(t\oplus \delta_t)||_2-||f_I(v)- f_T(t)||_2}{||f_I(v)- f_T(t)||_2}.
\end{equation}
We provide the distances averaged on 5000 image-text pairs from Flickr30K in Table \ref{tab:distance}. Benefiting from our delicate designs, C-PGC achieves better disruption of the aligned multimodal relationship, thereby boosting the generalization ability and transferability of the produced UAP.

\begin{table}[htbp]
  \centering
    \setlength{\tabcolsep}{5pt}
  \caption{Average relative cross-modal feature distances.}
    \resizebox{\linewidth}{!}{\begin{tabular}{cc|c|c|c|c|c|c} \toprule
    Source   & Method   & ALBEF    & TCL       & BLIP & X-VLM     & CLIP$_\textrm{ViT}$ & CLIP$_\textrm{CNN}$ \\ \midrule
    \multicolumn{1}{c}{\multirow{3}[0]{*}{ALBEF}} & GAP      & \gc7.18     & 6.54     & 0.91     & 1.74     & 0.31     & 0.98  \\ & ETU      & \gc8.70    &  8.02  &   0.81   & 2.85     &   0.36   & 1.32
    \\
             & C-PGC     & \gc\textbf{8.83} & \textbf{14.95} & \textbf{2.73} & \textbf{6.09} & \textbf{3.42} & \textbf{3.92} \\ \midrule
    \multirow{3}[0]{*}{TCL} & GAP      & 4.02     & \gc24.27    & 0.91     & 0.87     & 0.12     & 0.07  \\ & ETU      & 5.57     & \gc26.32    & 0.55     & 2.56     & 1.47     & 0.55 \\
             & C-PGC     & \textbf{6.43} & \gc\textbf{27.11} & \textbf{3.64} & \textbf{4.35} & \textbf{2.56} & \textbf{2.94} \\ \midrule
    \multirow{3}[0]{*}{BLIP} & GAP      & 3.17     & 4.67     & \gc11.82    & 1.74     & -1.71    & -0.98  \\ & ETU      & 5.26     & 6.03     & \gc13.14    & 2.90     & 0.25    & 1.14
    \\
             & C-PGC     & \textbf{6.41} & \textbf{12.15} & \gc\textbf{13.64} & \textbf{4.35} & \textbf{1.71} & \textbf{1.96} \\
             \bottomrule
    \end{tabular} } 
  \label{tab:distance}%
\end{table}

\section{Discussions and Future Directions}
\textbf{Overlook of interactions between perturbations $\delta_v$ and $\delta_t$}. The proposed framework generates universal perturbations for image and text respectively based on the designed loss functions. Despite the remarkable performance, it does not consider interactions between $\delta_v$ and $\delta_t$ during the optimization. 
% which has been leveraged in several previous attacks \citep{lu2023set, wang2024transferable} to improve performance. 
Future research can explore this limitation as a potential mechanism to further strengthen the attack.

{\textbf{Textual Semantic consistency.} To ensure the stealthiness of text attacks, we set the perturbation budget $\epsilon_t=1$, \ie, only one word is modified. Although C-PGC obtains superior semantic similarity to previous sample-specific methods, there is still room to improve from the proposed C-PGC. 
Moreover, future studies can consider similarity preservation strategies by applying more effective constraints during the generator training or adversarial sentences post-processing to facilitate a more stealthy attack.}

\section{More Visualization Results}
\begin{figure*}[t]
  \centering
  \includegraphics[width=\linewidth]{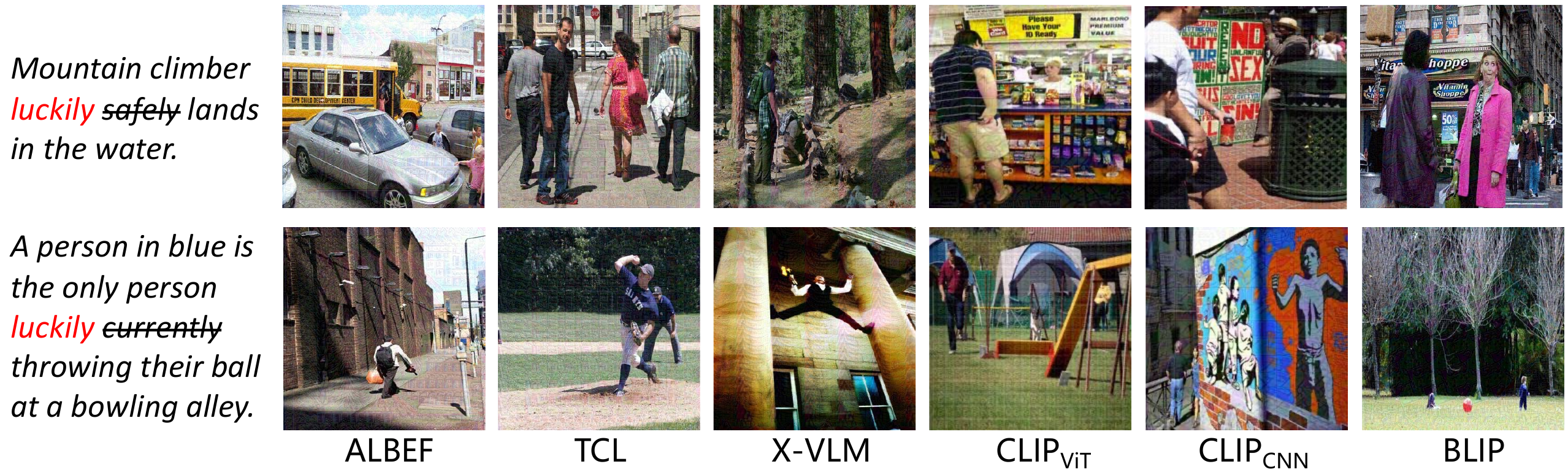}
  \caption{Attacks results of C-PGC on the image retrieval task. The red indicates the UAP and the crossed-out word is the replaced one. We generate the word on ALBEF and test it on 6 target models. All retrieved images fail to correspond to the query text, validating the rationality of our contrastive learning-based attacks.}
  % \vspace{-1em}
\label{fig:IR}
\vspace{1em}
 \end{figure*}

\begin{figure*}[t]
  \centering
  \includegraphics[width=\linewidth]{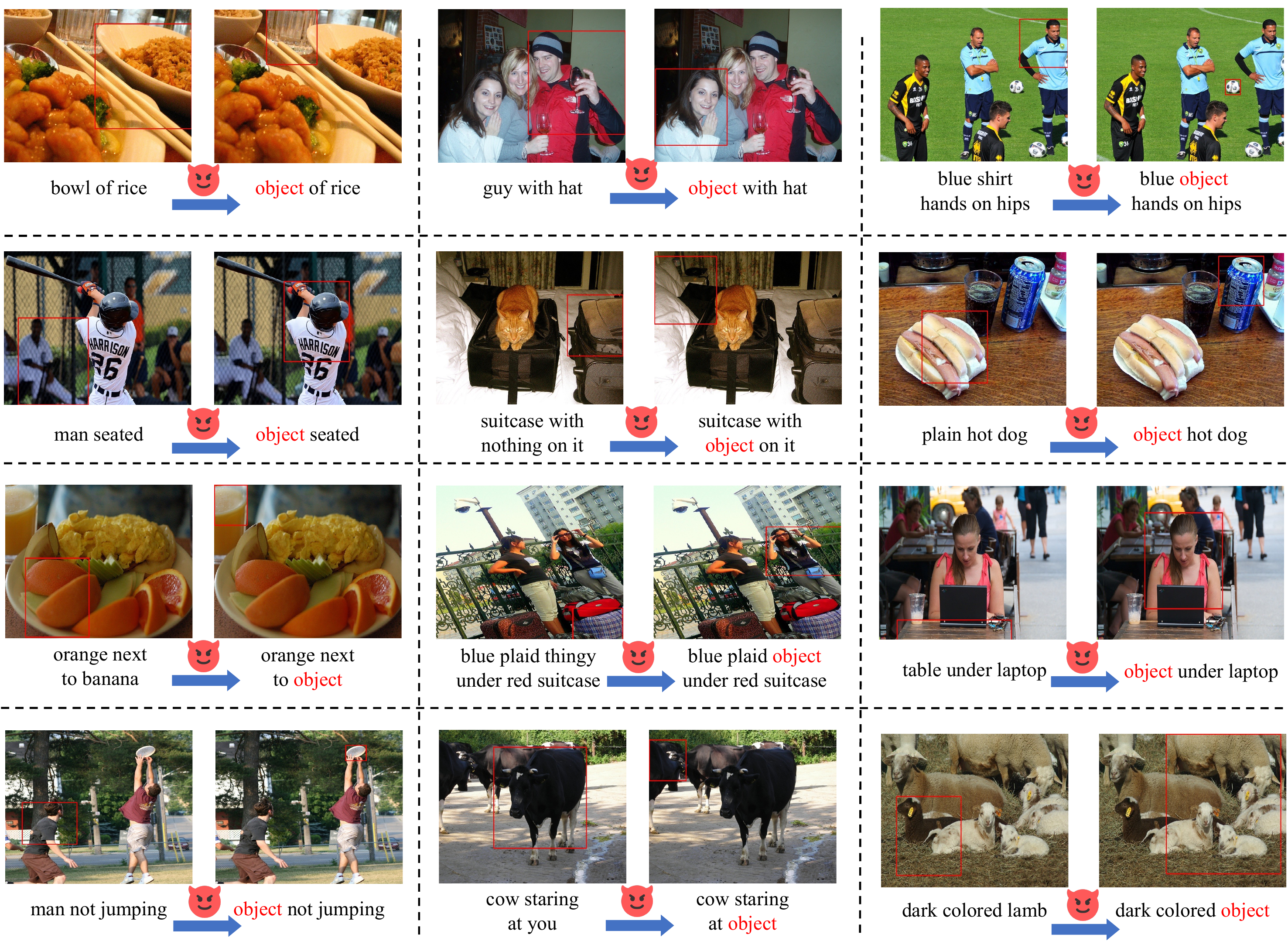}
  \centering
  % \vspace{-1em}
  \caption{Illustration of visual grounding. The predictions of clean pairs are on the left while the predictions of adversarial samples are on the right. The red word is the adversarial word perturbation.}
\label{fig:visualization_VG}
 \end{figure*}

\begin{figure*}[t]
  \centering
  \includegraphics[width=0.73\linewidth]{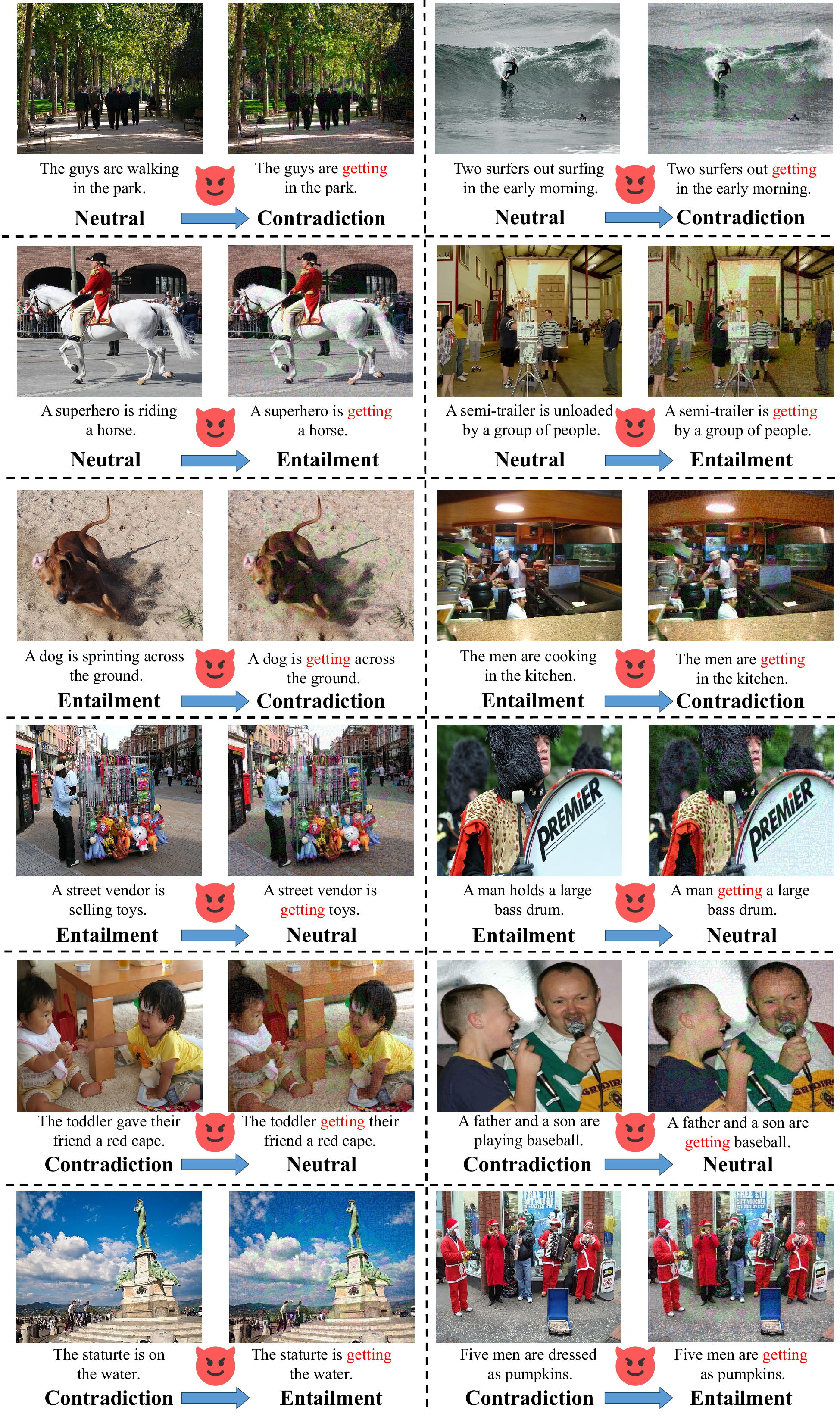}
  \centering
  \caption{Illustration of the visual entailment task. The red indicates the universal adversarial word. It can be observed that all predictions do not match with the ground truth.}
\label{fig:visualization_VE}
 \end{figure*}

This section presents a rich visual analysis of the proposed attack on a series of downstream tasks. Specifically, we first provide the visualization of the image retrieval task using the MSCOCO dataset in Fig. \ref{fig:IR}.
Besides, we generate the UAP using the ALBEF model for the visual grounding (VG) task. As illustrated in Fig. \ref{fig:visualization_VG}, the prediction bounding boxes exhibit a notable deviation from the clean predictions, verifying that our generated adversarial samples significantly interfere with the multimodal alignment. 
In the visual entailment (VE) task, we employ BLIP as the victim model and present the results in Fig. \ref{fig:visualization_VE}. These qualitative visualizations again demonstrate the remarkable attack effects of our proposed method on various downstream tasks.

 
\end{document}